%% file: main.tex
\documentclass[10pt,twocolumn,letterpaper]{article}

\usepackage[pagenumbers]{cvpr} 

\input{preamble}

\definecolor{cvprblue}{rgb}{0.21,0.49,0.74}
\usepackage[pagebackref,breaklinks,colorlinks,allcolors=cvprblue]{hyperref}

\def\paperID{232} 
\def\confName{CVPR}
\def\confYear{2026}

\title{SHOW3D: Capturing Scenes of 3D Hands and Objects in the Wild}

\author{%
\hspace*{-8pt}%
\begin{tabular}[t]{c}
Patrick Rim$^{1,2}$\thanks{}\quad\,
Kevin Harris$^{1}$\quad\,
Braden Copple$^{1}$\quad\,
Shangchen Han$^{1}$\quad\,
Xu Xie$^{1}$\quad\,
Ivan Shugurov$^{1}$\\
Sizhe An$^{1}$\quad\,
He Wen$^{1}$\quad\,
Alex Wong$^{2}$\quad\,
Tomas Hodan$^{1}$\quad\,
Kun He$^{1}$\\[0.2cm]
$^{1}$Meta Reality Labs\qquad
$^{2}$Yale University
\end{tabular}
}

\begin{document}

\twocolumn[{%
  \renewcommand\twocolumn[1][]{#1}%
  \maketitle
  \vspace{-25pt}
  \begin{center}
    \captionsetup{type=figure}
    \includegraphics[width=\textwidth]{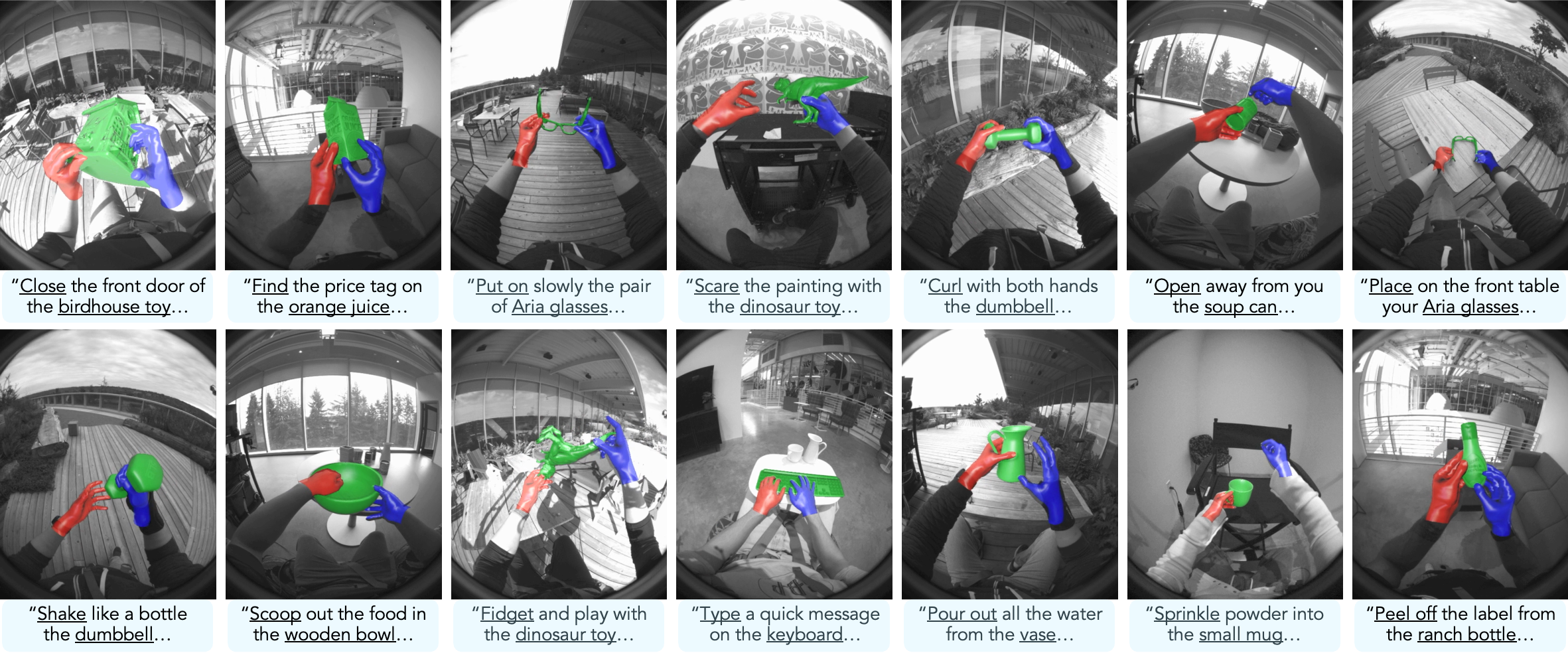}%
    \vspace{-6pt}
    \captionof{figure}{
    \textbf{SHOW3D} is the first dataset of in-the-wild hand–object interactions with accurate 3D annotations as well as text descriptions.
    The dataset was captured with our novel mobile multi-camera rig in diverse indoor and outdoor scenes, and annotated with 3D shapes and poses with our multi-view pipeline. Overlays show 3D annotations projected to egocentric images (hands in red and blue, object in green).}
    \label{fig:teaser}
  \end{center}%
  \vspace{4pt}
}]
\thispagestyle{plain}
{%
  \renewcommand{\thefootnote}{\fnsymbol{footnote}}%
  \footnotetext[1]{Work done during an internship at Meta.}%
}

\begin{center}
\section*{Abstract}
\end{center}
{\it
Accurate 3D understanding of human hands and objects during manipulation remains a significant challenge for egocentric computer vision.
Existing hand–object interaction datasets are predominantly captured in controlled studio settings, which limits both environmental diversity and the ability of models trained on such data to generalize to real-world scenarios. 
To address this challenge, we introduce a novel marker-less multi-camera system that allows for nearly unconstrained mobility in genuinely in-the-wild conditions, while still having the ability to generate precise 3D annotations of hands and objects.
The capture system consists of a lightweight, back-mounted, multi-camera rig that is synchronized and calibrated with a user-worn VR headset.
For 3D ground-truth annotation of hands and objects, we develop an ego-exo tracking pipeline and rigorously evaluate its quality.
Finally, we present SHOW3D, the first large-scale dataset with 3D annotations that show hands interacting with objects in diverse real-world environments, including outdoor settings.
Our approach significantly reduces the fundamental trade-off between environmental realism and accuracy of 3D annotations, which we validate with experiments on several downstream tasks. \url{show3d-dataset.github.io}
}

\vspace{10pt}
\section{Introduction}
\label{sec:intro}

\begin{figure*}[t]
  \centering
  \vspace{-1pt}
  \includegraphics[width=0.85\linewidth]{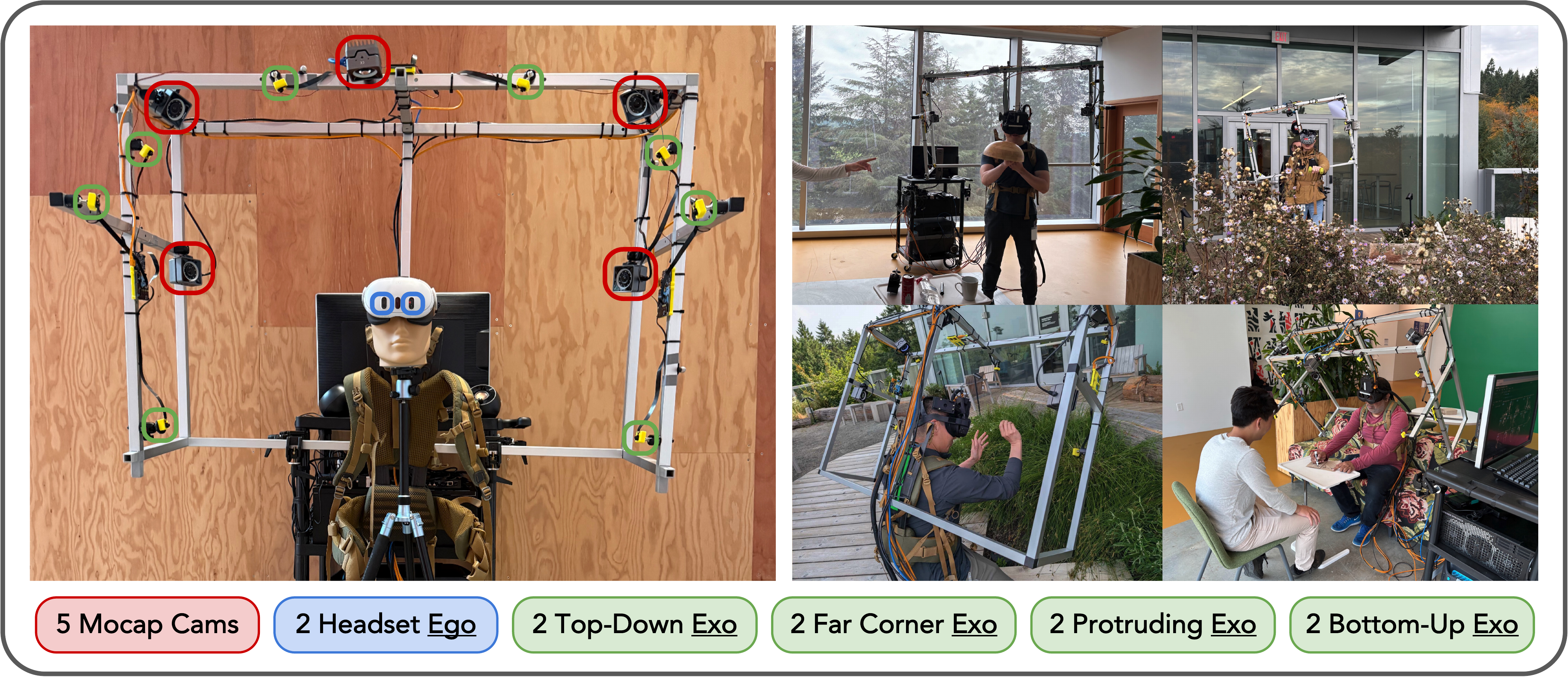}
  \vspace{-4pt}
  \caption{\textbf{Our mobile multi-camera capture rig.} Left: Annotated hardware layout showing five MoCap cameras (red), and eight exocentric monochrome cameras mounted in a half-dome configuration (green), and two egocentric monochrome cameras on the Meta Quest 3 headset (blue).  
  The MoCap cameras are only used for headset pose tracking, by tracking optical markers rigidly attached to the headset.
  The ten total exocentric and egocentric monochrome cameras are used for marker-free annotation of 3D hand and object poses. 
  Right: The rig in use during in-the-wild capture sessions, demonstrating its lightweight (about eight kilograms) and wearable design that allows natural interaction while maintaining stable, synchronized multi-view coverage under mostly unconstrained motion.}
  \label{fig:rig}
  \vspace{-2mm}
\end{figure*}

Accurate 3D tracking of human hands and their interactions with objects in everyday environments is essential for egocentric perception systems. 
In augmented and virtual reality (AR/VR), understanding of hands and manipulated objects enables natural and expressive user interactions without reliance on external input devices or constrained input spaces. 
In robotics and teleoperation, this ability is critical for advising and transferring dexterous manipulation skills. 

Despite recent advances, existing hand–object perception models~\cite{han2020megatrack, han2022umetrack, pavlakos2024reconstructing, zhang2016learning, yang2024mlphand, prakash20243d} often struggle when deployed outside controlled and indoor environments. 
There is therefore a pressing need for data collection systems that can produce realistic, unconstrained, and annotated captures of hand–object interactions \emph{in the wild}. 
However, building such systems is highly challenging due to a fundamental trade-off.
On one hand, the need for realistic hand/object appearance, environmental diversity, and user mobility limits the application of external sensors, \eg, cameras, IMUs, ToF, and motion capture systems~\cite{kim2009multi, sun20213drimr, xia2023quadric, alexiadis2016integrated, rim2025protodepth}. 
On the other hand, the availability and quality of 3D ground truth are bounded by sensor fidelity and coverage.

Existing hand–object interaction datasets often fall at one of two extremes in this trade-off. Recent datasets, including GigaHands~\cite{fu2025gigahands}, HOT3D~\cite{banerjee2025hot3d}, and ARCTIC~\cite{fan2023arctic}, are captured under controlled conditions, typically within indoor studio spaces equipped with motion capture systems or stationary multi-camera arrays. While these setups can offer high 3D accuracy, they inherently limit environmental diversity and, in the case of marker-based motion capture, hinder visual realism by altering natural hand and object appearance with reflective markers. At the other extreme, datasets with significant environmental variation, such as Ego-Exo4D~\cite{grauman2024ego} and Nymeria~\cite{ma2024nymeria}, lack dense or accurate 3D ground truth for hands and objects due to their limited sensor setup.

In this work, we break the trade-off between realism and ground-truth quality by designing a novel multi-camera capture system inspired by prior work~\cite{han2020megatrack} but
optimized for mobility. Our lightweight backpack-style rig acquires accurate 3D ground truth while enabling users to move freely, allowing for capture of realistic interactions in diverse environments, including the outdoors (Figure~\ref{fig:teaser}). 
The system provides ten fisheye monochrome cameras synchronized  at 60Hz, including eight exocentric ones attached on the rig and two egocentric ones from a user-worn Meta Quest 3 headset (Figure~\ref{fig:rig}).
To generate accurate ground-truth 3D poses of hands and objects, we develop a novel ego-exo annotation pipeline that leverages all ten camera views.
For hand pose annotation, we use state-of-the-art models for keypoint detection, and fuse them via robust triangulation with RANSAC.
For object pose annotation, we develop a multi-view extension of recent DINOv2-based methods for 2D detection~\cite{nguyen2023cnos}, pose estimation~\cite{ornek2024foundpose}, and refinement~\cite{nguyen2025gotrack}.
Both components produce their own confidence estimates, which allow for automated filtering when building training and evaluation datasets.
We assess the quality of both our hand and object pose annotations via evaluation against independent references, demonstrating sub-centimeter accuracy.

We use our novel capture system and annotation pipeline to collect \textbf{SHOW3D}, a large-scale dataset of hand–object interactions with dense and accurate 3D annotations, and the first such dataset that is captured across diverse real-world environments. Besides images from ten egocentric and exocentric cameras and 3D pose annotations of hands and objects, the dataset includes fitted 3D meshes, 2D segmentation masks, contact regions, and text descriptions.

Cross-dataset experiments show that models for 3D hand pose estimation and hand–object interaction field estimation trained on existing studio datasets generalize poorly to in-the-wild data, while SHOW3D-trained models maintain strong performance on studio benchmarks, confirming that our dataset captures broader environmental diversity. 
Furthermore, we demonstrate the utility of our text annotations through object trajectory forecasting experiments, where text-conditioned models significantly outperform baselines relying only on past trajectory, highlighting the value of semantic context for understanding hand–object interactions.

\vspace{1em}
\noindent In summary, we make the following contributions:
\begin{enumerate}
\item \textbf{A mobile data capture system:} The first capture setup for recording and automatically generating 3D annotations of in-the-wild hand–object interactions. Our setup consists of a mobile multi-camera capture rig that enables the acquisition of accurate ground-truth hand and object pose annotations from multi-view imagery.
\item \textbf{Ego-exo 3D ground truth:} We build a pipeline to annotate 3D hand and object poses using ego-exo fusion, and automatically filter the resulting annotations using reliable confidence estimates.
We also rigorously evaluate the quality of our 3D ground truth for hands and objects.
\item \textbf{Large and diverse dataset:} Our dataset SHOW3D comprises 4.3M frames of synchronized multi-view images with comprehensive annotations including 3D hand poses and MANO meshes, 6DoF object poses with 3D models, and descriptive action-level text captions.
\item \textbf{Demonstrated dataset utility:} Empirical validation via cross-dataset generalization experiments shows that models trained on SHOW3D transfer better to other datasets than vice versa, and text-conditioned forecasting experiments confirm the value of semantic annotations.
\end{enumerate}

\begin{figure*}[t]
  \centering
  \includegraphics[width=1.0\linewidth]{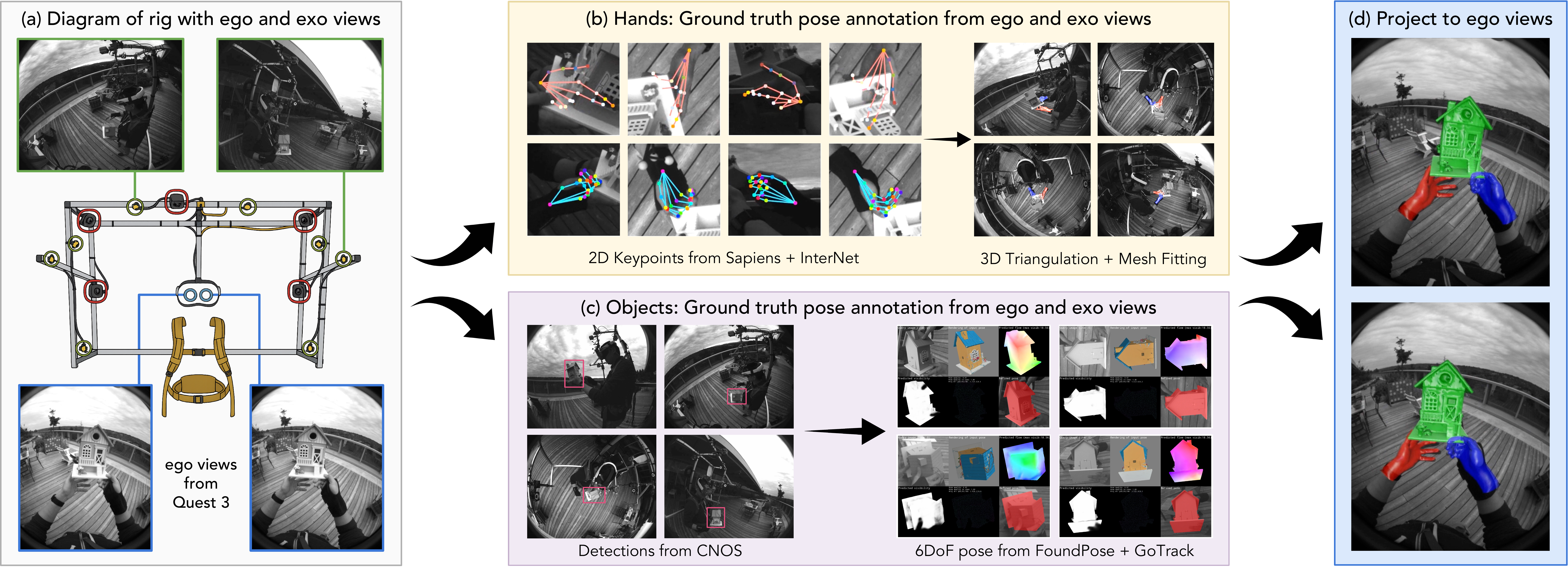}
  \caption{\textbf{Our ego-exo pipeline for 3D hand and object pose annotation.} 
  (a) Multi-view fisheye images from our ego and exo cameras. 
  (b) We detect 3D hand keypoints by fusing predictions from Sapiens~\cite{khirodkar2024sapiens} and InterNet~\cite{Moon_2020_ECCV_InterHand2.6M}, and fit personalized hand mesh via Inverse Kinematics. 
  (c) CAD-based 3D object pose estimation using CNOS~\cite{nguyen2023cnos}, FoundPose~\cite{ornek2024foundpose}, and GoTrack~\cite{nguyen2025gotrack}.
  (d) The resulting 3D ground-truth annotations are projected back into ego cameras, and can be used to train egocentric vision models.
  }
  \label{fig:pipeline}
\end{figure*}

\section{Related Work}
\label{sec:relatedwork}

In this section, we review the most recent efforts in collecting large-scale hand–object datasets. 
In particular, we emphasize the capture and annotation methodologies for \emph{real data}, as opposed to synthetic data~\cite{hewitt2024lookma,GANeratedHands_CVPR2018,li2023renderih}: while synthetic engines could alleviate the scarcity of high-quality data in certain scenarios, there generally remain significant challenges in simulating rich, realistic hand–object dynamics and environmental factors, often resulting in sim-to-real domain gaps that are difficult to overcome.

HOI4D~\cite{liu2022hoi4d} records RGB-D sequences using a Kinect v2 and an Intel RealSense camera mounted on a bicycle helmet worn by participants in indoor environments. Ground truth labels for hands and objects are obtained using manual annotation and frame-wise propagation. Despite its pioneering scale, relying solely on a single egocentric view severely constrains the fidelity of ground truth, as it is challenging to resolve occlusions and scale ambiguity.

Marker-based motion capture (MoCap) systems have been used extensively for ground truth capture purposes.
ARCTIC~\cite{fan2023arctic} uses 
an off-the-shelf solution with 54 infrared MoCap cameras to track hands and objects by attaching IR-reflective markers to them.
However, the stationary dome-style setup limits environmental diversity, and the presence of markers detracts from visual realism and impedes natural interactions.
OakInk2~\cite{zhan2024oakink2} employs a hybrid capture solution with 12 MoCap cameras and four RGB cameras (three exocentric, one egocentric). Similar to ARCTIC, both the hands and the objects are embedded with markers in order to generate 3D pose ground truth.
TACO~\cite{liu2024taco} uses a setup consisting of 12 static exocentric RGB cameras, one egocentric GoPro camera, and a MoCap system. 
While object poses are still marker-based, hand pose annotations are derived from a marker-less pipeline, using 2D keypoint detection and triangulation.
A recent dataset, HOT3D~\cite{banerjee2025hot3d}, works towards realism by utilizing the Meta Quest 3 headset and Aria research glasses \cite{engel2023projectarianewtool}.
However, the hands and objects are still tracked with a MoCap system relying on markers.

On the other hand, marker-less solutions typically rely on a synchronized multi-camera setup.
Assembly101~\cite{sener2022assembly101} and 
AssemblyHands~\cite{ohkawa2023assemblyhands} use 8 RGB cameras synchronized with a custom-built headset to capture hand–object interactions and annotate 3D hand poses, but lack ground-truth object pose. 
HO-Cap~\cite{wang2024hocapcapturedataset3d} uses multi-view RGB-D sensors to reconstruct 3D hand and object poses, and uses them as source motion in simulation.
More recently, GigaHands~\cite{fu2025gigahands} 
utilizes an array of 51 RGB cameras to build a 2D-to-3D pipeline for hands and objects.
The dataset notably offers a large number of frames compared to previous datasets,
but the stationary and restricted capture volume still results in  limited environmental and contextual diversity.
Furthermore, GigaHands does not contain egocentric views. 

Compared to static studios, mobile data capture systems are more difficult to build and operate.
A notable large-scale effort is Ego-Exo4D~\cite{grauman2024ego}, featuring one egocentric view and four exocentric views. However, since its primary focus is on capturing full-body actions in real-world scenarios, the cameras are not optimized for observing hands, and hand annotations are available only in a sparse subset of frames. Moreover, Ego-Exo4D does not provide 3D object annotations.
Nymeria~\cite{ma2024nymeria}, a similar real-world dataset captured with Aria research glasses and wearable MoCap suits, also focuses on full-body motion instead of detailed hand–object interaction. 
EgoDex~\cite{hoque2025egodexlearningdexterousmanipulation} is an egocentric dataset of dexterous manipulations collected using Apple Vision Pro API for hand pose estimation, but lacks object pose annotations, and is captured in simple studio-like environments.

Across all these existing efforts, an evident trade-off emerges: employing extensive multi-sensor capture systems can yield high-quality annotations, but sacrifices environmental diversity and visual realism.
To the best of our knowledge, we introduce the first truly mobile multi-camera capture rig capable of acquiring precise 3D hand–object ground truth for egocentric perception in genuinely unconstrained environments.
The closest work to ours is the back-mounted multi-camera system from MEgATrack~\cite{han2020megatrack}, but it is not designed to capture egocentric data.

\section{Data Collection and Annotation}

\subsection{Mobile Capture Rig}

We design a portable, self-contained, backpack-style multi-camera capture rig weighing roughly eight kilograms, which we found can be worn by human participants without significantly restricting their natural range of motion during a capture session.
Eight monochrome fisheye cameras ($1024\times1280$ resolution,  152$\degree$ horizontal and 116$\degree$ vertical field of view) are rigidly mounted on the rig in a half-dome configuration to maximize visual coverage.
See Figure~\ref{fig:rig} for the arrangement of cameras; exact camera placements and viewing angles were determined through empirical testing.

The participants also wear a Meta Quest 3 headset, and we record the egocentric images from the two front-facing monochrome fisheye cameras.
To allow for realistic head movements, we do not fix the headset to the rig.
Then, to track the headset pose relative to the capture rig, we employ MoCap the same way as in HOT3D~\cite{banerjee2025hot3d}: a lightweight marker tree (rigid body) is placed atop the headset, which is tracked via the five MoCap cameras on the rig.

All ten monochrome cameras and the MoCap cameras are hardware-synchronized at 60Hz, and precisely calibrated into a shared 3D reference frame. 
Note that the reference frame moves with the participant, instead of being fixed to the physical world.
Video and MoCap data are streamed to a high-performance desktop workstation placed on a mobile cart, which moves together with the participant guided by an operator, ensuring full mobility.

\begin{table*}[t]
\centering
\small
\renewcommand{\arraystretch}{1.0}

\resizebox{\textwidth}{!}{%
\begin{tabular}{l c c c c c c c c}
\toprule
Dataset & Ego / Exo & Frames & Hand Pose & Object Pose & Objects & Subjects & Environment & Annotation \\
\midrule
\rowcolor{gray!12}
\textbf{SHOW3D (ours)} & 2 / 8 & 4.3M  & Both / Mesh & \cmark & 21 & 38 & In-the-wild & Monochrome + optimization \\
GigaHands~\cite{fu2025gigahands} & 0 / 51 & 3.7M  & Both / Mesh & \cmark & 417 & 51& Studio & RGB + optimization \\
ARCTIC~\cite{fan2023arctic} & 1 / 8 & 2.1M  & Both / Mesh & \cmark & 11 & 10 & Studio & MoCap \\
HOT3D~\cite{banerjee2025hot3d} & 3 / 0 & 1.7M & Both / Mesh & \cmark & 33 & 19 & Studio & MoCap \\
TACO~\cite{liu2024taco} & 1 / 12 & 363K  & Both / Mesh & \cmark & 196 & 14 & Studio & MoCap \\
OakInk2~\cite{zhan2024oakink2} & 1 / 3 & 993K & Both / Mesh & \cmark & 75 & 9 & Studio & MoCap \\
HOI4D~\cite{liu2022hoi4d} & 2 / 0 & 2.4M & Both / Mesh & \cmark & 800 & 4 & Studio & RGB-D + manual \\
HO-Cap~\cite{wang2024hocapcapturedataset3d} & 1 / 9 & 699K & Both / Mesh & \cmark & 64 & 9 & Studio & RGB-D + optimization \\
HOGraspNet~\cite{cho2024dense} & 0 / 4 & 1.5M & Single / Mesh & \cmark & 30 & 99 & Studio & RGB-D + optimization \\
\midrule
Ego-Exo4D~\cite{grauman2024ego} & 2 / 4 & $>$100M & Both / Skel. & \xmark & Many & 740 & In-the-wild & RGB + manual \\
AssemblyHands~\cite{ohkawa2023assemblyhands} & 4 / 8 & 203K & Both / Skel. & \xmark & Many & 20 & Studio & RGB + manual \\
EgoDex~\cite{hoque2025egodexlearningdexterousmanipulation} & 1 / 0 & 90M & Both / Skel. & \xmark & Many & -- & Studio-like & Apple Vision Pro API \\
\bottomrule
\end{tabular}
} 
\vspace{-1mm}
\caption{\textbf{Comparison of SHOW3D against recent hand–object interaction datasets.}}
\label{tab:dataset_comparison}
\end{table*}

\subsection{3D Hand Pose Annotation}
\label{sec:hand_pose}
We design an ego-exo fusion pipeline to obtain 3D hand poses from all ten  fisheye cameras in our system.
First, we apply Sapiens~\cite{khirodkar2024sapiens}, a state-of-the-art human pose  estimation model, to each camera view and obtain body keypoint detections, including 21 keypoints per hand. 
We use a version of Sapiens finetuned on proprietary human keypoint datasets, including extensive amounts of synthetic data that, in aggregate, can approximate aspects of real-world conditions encountered during in-the-wild captures.

Given the 2D hand keypoint predictions from Sapiens, we next triangulate them into the 3D space defined by the capture rig.
However, it is non-trivial to directly triangulate with fisheye distortion.
To simplify the problem, we first perform perspective cropping~\cite{han2022umetrack} around the region of detected hand keypoints within each camera view, and warp the keypoint detections to the resulting virtual pinhole cameras.

In addition to the Sapiens keypoints, we detect another set of 2D hand keypoints from the perspective crops, using a dedicated hand pose estimation model, InterNet~\cite{Moon_2020_ECCV_InterHand2.6M}, which we also finetuned on additional studio and synthetic datasets.
The reasoning is that detecting hand keypoints on cropped images is a simpler problem and can have higher accuracy, whereas directly detecting them from the full image can be suboptimal due to insufficient feature resolution for hands in the Sapiens model.
We fuse the two sets of 2D hand keypoint detections by performing RANSAC-based triangulation with both sets.
We find that including hand crops from the egocentric cameras often helps performance, as they capture hands from angles that are distinct from the exocentric views.

Next, the triangulated 3D hand keypoints are used to fit detailed hand meshes. 
We use a solution compatible with UmeTrack~\cite{han2022umetrack} and HOT3D~\cite{banerjee2025hot3d}, where a personalized linear blend skinning model of the hand is first derived from a high-resolution hand scan system;
this personalization ensures high-quality ground truth meshes suitable for fine-grained tasks, such as contact estimation.
Then, the personalized model is fit to the 3D keypoints via Inverse Kinematics, to obtain a temporally smooth sequence of hand poses.
Finally, we estimate hand confidence using a Bayesian formulation comprising two terms: one based on the errors in keypoint detection and triangulation, and the other based on the residual error in Inverse Kinematics.

\subsection{3D Object Pose Annotation}
To generate accurate 6DoF object pose annotations, we develop a CAD-based pipeline consisting of three stages: (1)~2D object detection by CNOS~\cite{nguyen2023cnos}, (2)~coarse 6DoF object pose estimation by FoundPose~\cite{ornek2024foundpose}, and (3)~6DoF object pose refinement by the GoTrack refiner~\cite{nguyen2025gotrack}.

We started with open-source implementations of these methods and extended them to support multi-view input images, which was motivated by the accuracy gains achieved with a multi-view extension of FoundPose presented in~\cite{banerjee2025hot3d}. 
The 2D object detector is run on all available views, and the detection with the highest confidence in each view is assumed to correspond to the target object (at most one instance of each object class is assumed present in the scene). The support of calibrated multi-view images in FoundPose and GoTrack is realized by replacing standard PnP with multi-view gPnP from PoseLib~\cite{PoseLib}. Besides boosting the pose accuracy, we observed that the multi-view input makes final confidence scores produced by the GoTrack refiner more reliable, which is an important property for filtering the generated pose annotations.

The first two stages are executed only in the first frame of a sequence or if the confidence score of the pose estimate from the previous frame is below a threshold. If the confidence score is sufficiently high, we execute only the pose refiner, using the previous pose estimate as the initial pose. This increases efficiency and also robustness in cases where object occlusion is gradually increasing.

All three stages are based on DINOv2 features~\cite{oquab2023dinov2} that allow for effective generalization to various image and camera types. In addition, none of the stages require any object-specific training, thanks to which the pipeline can be promptly applied to any object with an available CAD model. We use a subset of the objects from HOT3D~\cite{banerjee2025hot3d} for which high-quality CAD models are available.

\subsection{Instructions-to-Text Captions}
Inspired by GigaHands~\cite{fu2025gigahands}, each recording session in our dataset is guided by a structured, goal-oriented protocol designed to elicit diverse, natural hand–object interactions from participants. 
Participants are instructed to perform common object manipulations, such as ``spraying with a bottle’’, ``sipping from a mug’’, or ``cleaning a keyboard’’. Our instructions are intentionally broad and goal-oriented, promoting natural diversity and spontaneity.

To create rich natural language annotations from these instructions, we employ an LLM-based~\cite{anthropic_claude_2024} augmentation pipeline that generates multiple semantically diverse paraphrases for each interaction. 
This process yields a varied collection of text captions for every captured clip, providing valuable paired natural language descriptions that enhance the dataset's utility for language-conditioned models and multimodal training.

\section{The SHOW3D Dataset}
Using our mobile capture system, we collect SHOW3D, a large-scale dataset comprising over 4 million frames captured from 30 subjects performing natural manipulations with 21 everyday objects. Each frame is simultaneously captured by all ten synchronized cameras, providing comprehensive multi-view coverage. The dataset spans genuinely diverse environments, including a wide range of indoor and outdoor settings that are largely absent from existing hand–object interaction datasets.

\customparagraph{Indoor and outdoor scenes.} Our outdoor captures include garden areas with natural vegetation, outdoor seating areas with furniture and architectural elements, and even recordings from building ledges where the background is predominantly sky. These outdoor environments exhibit naturally varying illumination conditions due to weather changes, ranging from bright sunny days to overcast skies, along with dynamic shadows cast by overhead structures, trees, and varying sun angles. Indoor captures encompass kitchens, meeting rooms, hallways, cafeterias, desk areas, and recreational spaces with games and equipment. Indoor settings feature substantial illumination variability due to different artificial lighting configurations—from dim ambient lighting to bright overhead fluorescents—as well as shadow patterns from furniture and architectural features. This environmental and lighting diversity presents challenges rarely seen in studio datasets, where controlled conditions typically maintain consistent illumination and backgrounds.

\customparagraph{Annotations.} Our annotation pipeline produces rich ground-truth information: 3D hand poses for both hands (21 keypoints per hand with fitted personalized meshes), 6DoF object poses with texture-mapped 3D CAD models, 2D segmentation masks following~\cite{ravi2024sam}, and hand–object contact annotations following~\cite{taheri2020grab}. Multiple natural language captions generated through our LLM-based~\cite{anthropic_claude_2024} augmentation pipeline accompany each interaction sequence, offering semantic descriptions at varied levels of specificity. 

\customparagraph{Annotation quality.} SHOW3D is the \emph{first} hand–object interaction dataset to quantitatively evaluate both its 3D hand pose and 6DoF object pose annotations against independent references. 
To assess the accuracy of our GT \emph{hand} pose annotations, we compare the 3D hand keypoints from our system against gold-standard annotations obtained from two sources: automatic annotation from a large multi-camera studio, and manual annotation on data captured in-the-wild.
Across all conditions, our 3D keypoints achieve sub-centimeter median accuracy, approaching that of larger multi-camera systems.
To assess the accuracy of our GT \emph{object} pose annotations, we attach optical markers to a set of selected objects and track them using a MoCap system while they are being manipulated by hands.
We evaluate our GT object poses against these gold-standard MoCap object poses and find sub-centimeter P50 translation error across all objects, validating the quality of our annotations.
We present full experimental results in the supplementary material.

\begin{figure}[t]
  \centering
  \includegraphics[width=1.0\linewidth]{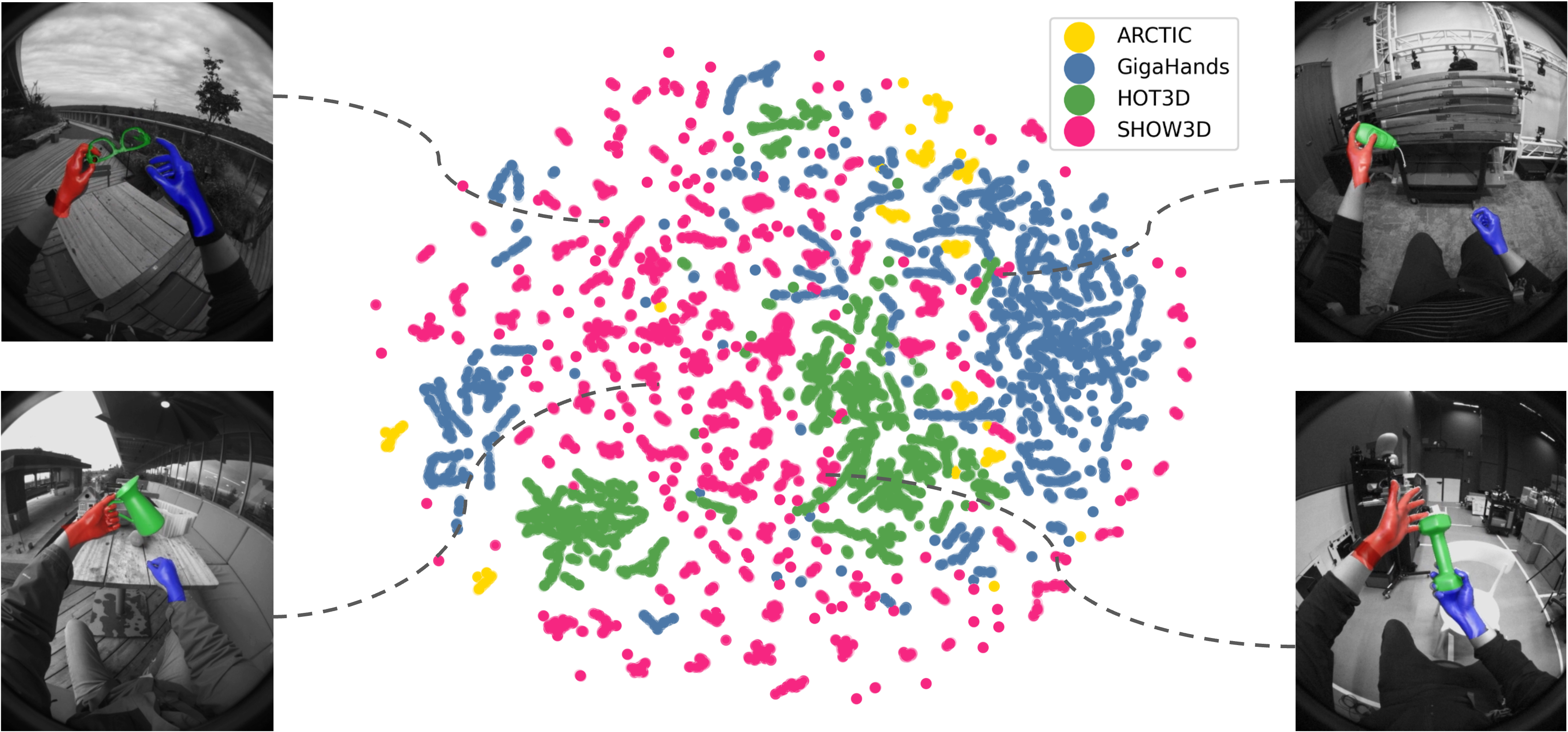}
  \caption{\textbf{Cross-dataset feature embedding.} 
  We plot UMAP~\cite{mcinnes2018umap-software} embeddings of DINOv2~\cite{oquab2023dinov2} features extracted from the raw images across different hand–object interaction datasets. 
  SHOW3D (pink) spans diverse visual domains between datasets collected in controlled environments: GigaHands (blue), HOT3D (green), ARCTIC (yellow).
  Best viewed in color.
  }
  \label{fig:umap}
  \vspace{-6pt}
\end{figure}

\customparagraph{Diversity.} As illustrated in Figure~\ref{fig:umap}, SHOW3D occupies a broad manifold in the feature space of hand–object interactions extracted from a pretrained backbone, bridging the compact clusters formed by studio datasets such as HOT3D, GigaHands, and ARCTIC. Frames that appear near these clusters visually correspond to similar conditions (\eg, one SHOW3D scene near the HOT3D cluster was captured in a comparable indoor lab setting, while another near the GigaHands cluster features trusses resembling those that host cameras in GigaHands). This shows that SHOW3D spans both in-the-wild and controlled domains, reflecting genuine environmental overlap while maintaining broader coverage.

\customparagraph{Comparison with existing datasets.} Table~\ref{tab:dataset_comparison} positions SHOW3D relative to existing datasets. While studio-based datasets achieve high annotation quality through controlled setups, they sacrifice environmental realism and diversity. Conversely, large-scale in-the-wild datasets like Ego-Exo4D capture authentic interactions but provide only sparse or manual 3D annotations without object pose. SHOW3D uniquely combines both: our mobile rig and RGB-based optimization pipeline enable dense, accurate 3D annotations of hands and objects across truly unconstrained real-world conditions, filling a critical gap in the hand–object interaction literature.

\section{Experiments}

\subsection{Hand--Object Interaction Field Estimation}

Existing contact detection methods primarily focus on binary contact estimation, determining whether hands and objects are in contact. However, in dexterous bimanual interactions, hands are often near objects without making contact. To capture richer spatial relationships, we adopt the \emph{interaction field estimation} task introduced in~\cite{fan2023arctic} that models the continuous spatial proximity between hands and objects throughout the course of an interaction.

\begin{table}[b]
\centering
\resizebox{\columnwidth}{!}{%
\begin{tabular}{cccc}
\toprule
Train set & Test set & ADE (mm) $\downarrow$ & ACC (m/s$^2$) $\downarrow$ \\
\midrule
SHOW3D & HOT3D & 14.70 & 4.05 \\
HOT3D & HOT3D & 11.29 & 3.21 \\
HOT3D + SHOW3D & HOT3D & \textbf{8.80} & \textbf{2.16} \\
\midrule
HOT3D & SHOW3D & 22.57 & 5.61 \\
SHOW3D & SHOW3D & 13.82 & \textbf{3.79} \\
SHOW3D + HOT3D & SHOW3D & \textbf{13.50} & 3.84 \\
\bottomrule
\end{tabular}
}
\caption{\textbf{Evaluation of interaction field estimation.}}
\label{tab:cross_field_estimation}
\end{table}

\begin{table*}[t]
\centering
\resizebox{\textwidth}{!}{%
\begin{tabular}{c l *{13}{c} c}
\toprule
\shortstack{Frames\\ahead} &  &
\rotatebox{45}{aria} & \rotatebox{45}{bowl} & \rotatebox{45}{cansoup} &
\rotatebox{45}{dinotoy} & \rotatebox{45}{dumbbell} & \rotatebox{45}{juice} & \rotatebox{45}{keyboard} &
\rotatebox{45}{milk} & \rotatebox{45}{mouse} & \rotatebox{45}{mug} &
\rotatebox{45}{mustard} & \rotatebox{45}{vase} &
\rotatebox{45}{waffles} &
Mean \\
\midrule

\multirow{2}{*}{30} & w/o text
 & 58.5 & 43.1 & 28.3 & 45.7 & 89.7 & 30.3 & 51.0 & 31.9 & 26.6 & 32.6 & 57.5 & 24.2 & 36.1
 & 42.7 \\
 & w/ text
 & \textbf{49.0} & \textbf{37.5} & \textbf{25.0} & \textbf{34.0} & \textbf{73.8} & \textbf{19.3} & \textbf{37.3} & \textbf{19.5} & \textbf{14.5} & \textbf{21.6} & \textbf{16.3} & \textbf{20.3} & \textbf{27.0}
 & \textbf{30.4} \\
\midrule

\multirow{2}{*}{60} & w/o text
 & 63.6 & 40.5 & 29.0 & 46.4 & 95.9 & 33.1 & 66.5 & \textbf{22.6} & 27.5 & 37.4 & 77.7 & 27.3 & 39.9
 & 46.7 \\
 & w/ text
 & \textbf{57.0} & \textbf{32.7} & \textbf{28.4} & \textbf{37.5} & \textbf{89.0} & \textbf{21.1} & \textbf{34.6} & 25.2 & \textbf{24.7} & \textbf{24.4} & \textbf{19.0} & \textbf{26.3} & \textbf{35.1}
 & \textbf{35.0} \\
\bottomrule
\end{tabular}%
}
\caption{\textbf{Evaluation of text-driven 6DoF object pose forecasting.} Reported are per-object and mean average translation errors (mm, $\downarrow$), with and without text conditioning.
Conditioning on text leads to a consistent improvement across objects.
}
\label{tab:show3d_text_vs_notext_mean}
\end{table*}

\customparagraph{Problem formulation.} For each vertex of each hand we aim to infer the closest distance to the object and vice versa. Specifically, an interaction field $\mathbf{F}^{a \rightarrow b} \in \mathbb{R}^{V_a}$ is defined as the distance to the closest vertex on mesh $\mathbf{M}_b$ for all vertices in mesh $\mathbf{M}_a$. $V_a$ denotes the number of vertices in mesh $\mathbf{M}_a$ (the reverse interaction field would be $\in \mathbb{R}^{V_b}$). We can formulate this problem mathematically, as follows:
\begin{equation}
\mathbf{F}^{a \rightarrow b}_i = \min_{1 \leq j \leq V_b} \left\lVert \mathbf{v}^a_i - \mathbf{v}^b_j \right\rVert_2, 
\qquad 1 \leq i \leq V_a,
\end{equation}
where $\mathbf{v}^{(\cdot)}_k \in \mathbb{R}^3$ represents the $k$-th vertex of mesh $\mathbf{M}_{(\cdot)}$. Our task is to estimate the interaction fields $\mathbf{F}^{l \rightarrow o}$, $\mathbf{F}^{r \rightarrow o}$, $\mathbf{F}^{o \rightarrow l}$, and $\mathbf{F}^{o \rightarrow r}$ for each frame, where $l$, $r$, and $o$ denote the left hand, right hand, and object meshes, respectively.

\customparagraph{Experimental setup.} We train the InterField~\cite{fan2023arctic} model and evaluate cross-dataset generalization between SHOW3D and HOT3D. We report two metrics: Average Distance Error (ADE) in millimeters, which measures the mean absolute error in predicted distances, and Acceleration (ACC) in $\text{m/s}^2$, which captures temporal smoothness of predictions.

\customparagraph{Results.} Table~\ref{tab:cross_field_estimation} presents cross-dataset evaluation results. When testing on HOT3D, training on SHOW3D alone achieves 14.70\,mm ADE compared to 22.57\,mm (+54\%) when training on HOT3D and testing on SHOW3D. This asymmetry demonstrates that models trained on SHOW3D generalize better to studio environments than vice versa, indicating that our in-the-wild data captures a broader distribution of interaction patterns and environmental conditions.

Furthermore, when using both datasets for training, we observe substantial improvements on HOT3D (8.80 vs.\ 11.29\,mm, a 22\% improvement) but minimal gains for SHOW3D evaluation (13.50 vs.\ 13.82\,mm, only 2\% improvement). This reveals that the domain of SHOW3D substantially encompasses the more constrained studio setting of HOT3D, while HOT3D provides limited additional coverage beyond what SHOW3D already captures. These results confirm that the environmental diversity in SHOW3D directly translates into improved model generalization, validating the value of in-the-wild hand--object interaction data.

\subsection{Text-driven 6DoF Object Pose Forecasting}

To validate the utility of our semantic text annotations, we evaluate whether natural language descriptions of hand-object interactions provide valuable context for predicting future object poses. Given a sequence of observed object poses and an optional text caption describing the interaction, the task is to forecast the 6DoF object pose at a future time step.
Natural language captures high-level interaction intent, such as ``pouring from a bottle’’ versus ``placing a bottle down’’, which helps disambiguate similar pose trajectories that lead to different future outcomes.

\customparagraph{Experimental setup.} We implement a transformer-based forecasting model that takes as input a sequence of past object poses, where each pose is represented as a 7-dimensional vector (3D position and quaternion for 3D rotation). The pose sequence is embedded and augmented with sinusoidal positional encodings before being processed by a multi-layer transformer encoder. For text conditioning, we encode captions using BERT~\cite{devlin2019bert} and project the text features into the same latent space as the pose embeddings. Text and pose features are concatenated and processed jointly by the transformer. The model predicts future object pose using separate heads for position (3D translation) and rotation (normalized quaternion). 
We train models both with and without text conditioning, using sequences of 30 past poses to predict the pose 30 or 60 frames into the future. 
We evaluate on sequences from our test set and report average translation error in millimeters across 13 objects.

\customparagraph{Results.} Table~\ref{tab:show3d_text_vs_notext_mean} presents per-object and mean translation errors for both forecasting horizons. At 30 frames ahead, text conditioning reduces mean error from 42.7 to 30.4mm, a 29\% improvement. This gain is consistent across nearly all objects, with particularly large improvements for objects like mug (34\%) and mustard (72\%). At the longer 60-frame horizon, text conditioning maintains its advantage, reducing mean error from 46.7 to 35.0mm (25\%).

These results demonstrate that semantic descriptions provide meaningful contextual information that complements purely kinematic patterns.
The consistent improvements across objects and time horizons validate the value of our text annotation pipeline and highlight the potential of language-conditioned models for understanding and predicting hand–object interactions.
Ultimately, SHOW3D's multimodal annotations enable richer downstream applications beyond purely visual perception.

\subsection{3D Hand Pose Estimation}

\begin{table}[t]
\centering
\resizebox{\columnwidth}{!}{%
\begin{tabular}{rccl}
\toprule
\# & Training set & Test set & MKPE (mm) $\downarrow$ \\
\midrule
1 & UmeTrack & SHOW3D & 22.2 (+55\%) \\
2 & HOT3D & SHOW3D & 19.6 (+37\%) \\
3 & UmeTrack + HOT3D  & SHOW3D & 16.4 (+15\%)\\
4 & SHOW3D  & SHOW3D & 15.5 (+8\%)\\
5 & UmeTrack + HOT3D + SHOW3D & SHOW3D & \textbf{14.3} \\
\midrule
6 & HOT3D & HOT3D & 14.0 (+14\%) \\
7 & UmeTrack + HOT3D  & HOT3D & 12.7 (+3\%) \\
8 & UmeTrack + HOT3D + SHOW3D & HOT3D & \textbf{12.3} \\
\midrule
9 & UmeTrack & UmeTrack & \phantom{0}9.7 (+2\%) \\
10 & UmeTrack + HOT3D  & UmeTrack & \phantom{0}\textbf{9.5} \\
11 & UmeTrack + HOT3D + SHOW3D & UmeTrack & \phantom{0}9.6 (+1\%) \\
\bottomrule
\end{tabular}
}%
\caption{\textbf{Evaluation of 3D hand pose estimation.} The reported metric is Mean Keypoint Position Error (MKPE, in mm). Models trained on existing datasets, UmeTrack~\cite{han2022umetrack} and HOT3D~\cite{banerjee2025hot3d}, generalize noticeably worse to SHOW3D, highlighting the increased difficulty of in-the-wild data.
}
\label{tab:cross_hands}
\end{table}

\customparagraph{Experimental setup.}
We evaluate the utility of our dataset as a benchmark for 3D hand pose estimation. 
We sample four subjects (with gender balance) from SHOW3D to build a balanced evaluation set, and apply a threshold on the estimated hand pose confidence described in Sec.~\ref{sec:hand_pose} to the entire dataset to filter valid frames for training and evaluation.
We benchmark generalization by training the multi-view UmeTrack hand tracking model~\cite{han2022umetrack} on 
SHOW3D, as well as evaluating cross-dataset generalization on the existing UmeTrack and HOT3D datasets.

Since we rely on marker-less ground truth generation in genuinely challenging in-the-wild conditions, the yield rate of training frames (\ie frames with valid hand pose annotation) in SHOW3D---using automated filtering---is naturally less than perfect.
Overall, SHOW3D retains 1.9M high-confidence training frames for hand pose estimation. Notably, this is greater than either UmeTrack or HOT3D, both of which have fewer than 1.4M training frames.

\customparagraph{Results.} 
As shown in Table~\ref{tab:cross_hands}, models trained on UmeTrack or HOT3D can achieve reasonable performance when tested within their respective domains (rows 6 and 9), but regress substantially when evaluated on SHOW3D (rows 1 and 2). 
A model trained on SHOW3D (row 4) achieves a significantly lower error on SHOW3D than a model trained on UmeTrack, HOT3D, or their combination (rows 1, 2, and 3).
We also observe that combining SHOW3D with the other datasets is consistently beneficial. 
The model trained on the combination of UmeTrack, HOT3D, and SHOW3D outperforms other models when evaluated on SHOW3D and HOT3D (rows 5 and 8), and is a close second on UmeTrack (row 10), which was captured in a very restricted environment.

\begin{figure}[t]
  \centering
  \includegraphics[width=1.0\linewidth]{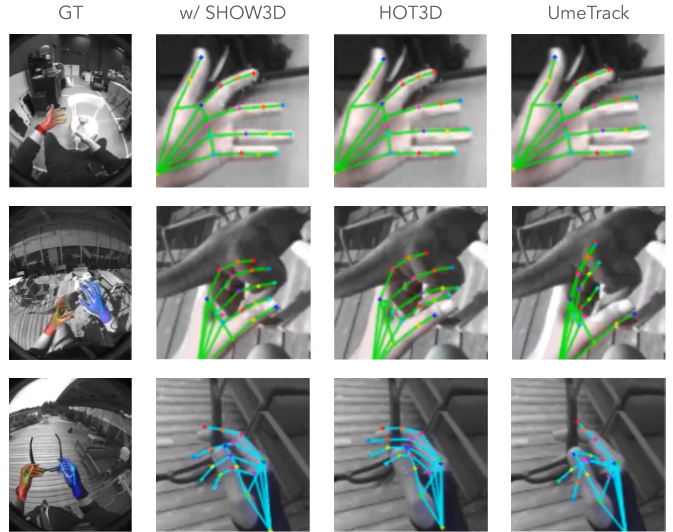}
  \caption{\textbf{Hand pose estimation on the SHOW3D test set.} 
  From left to right: results from models trained on UmeTrack + HOT3D + SHOW3D, HOT3D, and UmeTrack, respectively.
  Including training data from SHOW3D significantly improves robustness against object occlusion and background clutter.
  }
  \label{fig:hand_pose}
\end{figure}

Figure~\ref{fig:hand_pose} shows qualitative examples on the SHOW3D test set.
First, due to the lack of object interactions in the UmeTrack training set, the UmeTrack-trained model struggles in the face of even minor occlusions from objects.
We observe that the HOT3D-trained model handles occlusions better, but still achieves suboptimal performance, especially in environments with novel background/lighting.
Finally, the model trained with SHOW3D data shows significantly improved robustness to object interaction and background clutter.
Once again, these results highlight the unique value of real in-the-wild data with 3D ground truth, which offers robustness benefits not captured by prior datasets collected in more controlled environments (indoor studios).

\section{Conclusion}
We present SHOW3D, the first in-the-wild dataset with 3D ground truth for egocentric perception of hand–object interaction. To capture real-world diversity, SHOW3D features over 1,200 video sequences collected across more than 30 distinct locations.
SHOW3D is captured with a lightweight multi-camera rig optimized for high mobility, and annotated by an automated ego-exo pipeline that produces accurate 3D hand and object poses.
It is demonstrated on three downstream tasks that SHOW3D enables learning robust and generalizable models, proving its value over existing datasets that are predominantly captured in restricted environments.

While this work has focused on the visual understanding of hand–object interaction, future work will explore the integration of additional modalities, such as tactile sensing and range.
Furthermore, we are interested in expanding the downstream use cases of our data collection capabilities in embodied applications, such as teleoperation for robotics.

{
    \small
    \bibliographystyle{ieeenat_fullname}
    \bibliography{main}
}

\clearpage   
\appendix

\twocolumn[{%
  \renewcommand\twocolumn[1][]{#1}%
\maketitlesupplementary
  \vspace{-8pt}
  \begin{center}
    \captionsetup{type=figure}
    \includegraphics[width=.82\textwidth]{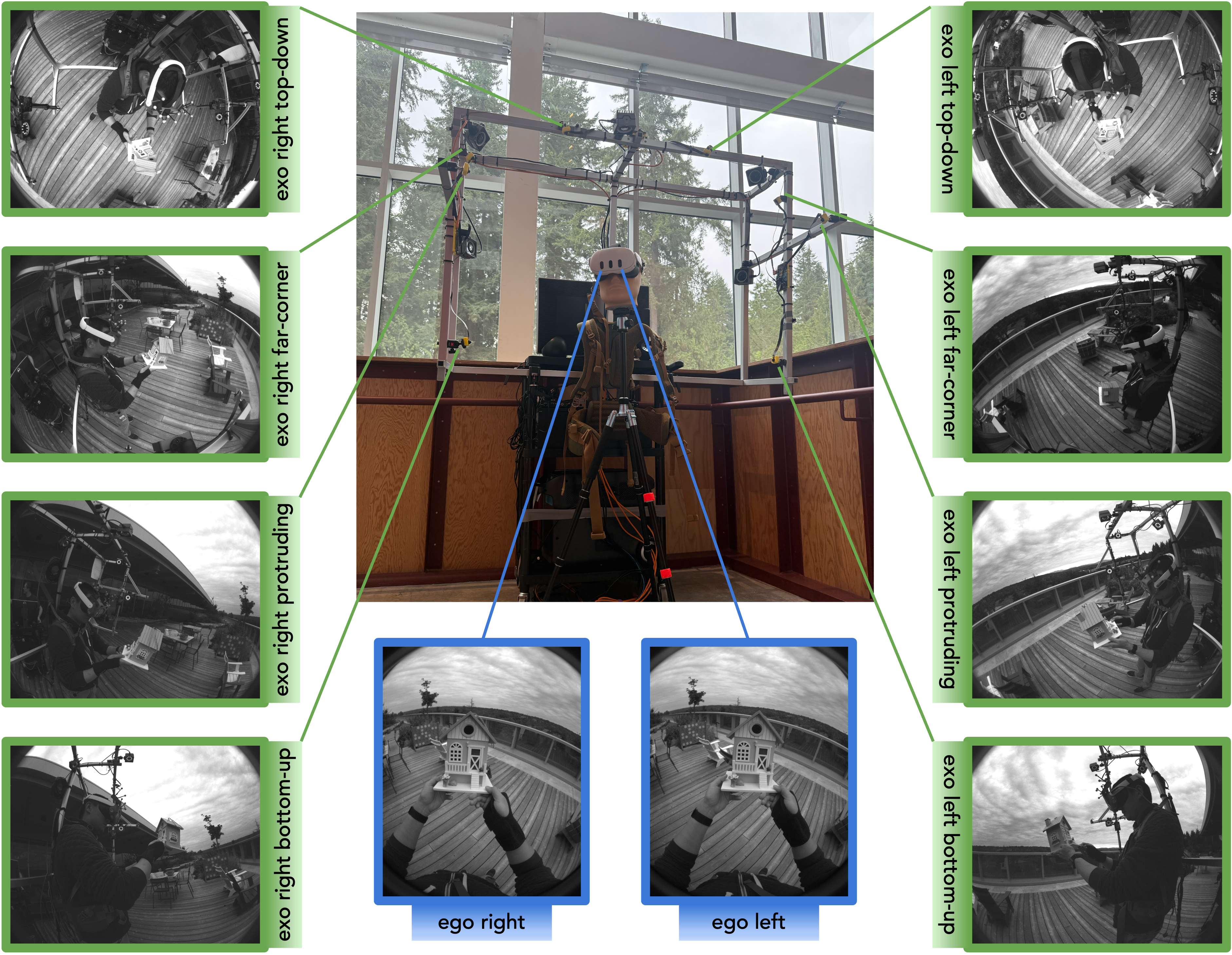}%
    \captionof{figure}{
    \textbf{SHOW3D mobile capture setup.} The rig includes eight exocentric fisheye cameras (green) providing wide-baseline views from diverse angles, and two egocentric fisheye cameras (blue) mounted on the headset. This configuration enables full-surround imaging of hand–object interactions. We use monochrome cameras instead of RGB as they are more robust to illumination variability.
    }
    \label{fig:rig_details}
  \end{center}%
  \vspace{5pt}
}]

\section{Mobile Capture Rig Details}
Figure~\ref{fig:rig_details} depicts our mobile capture rig and example images captured by each camera.
All camera and MoCap data are streamed, via USB and Ethernet, to 
a high-performance desktop workstation placed on a rolling cart.
The specs of the workstation are:
AMD Ryzen Threadripper PRO 3995WX 64-core processor, 128GB RAM, and 30TB SSD storage.
We power the workstation with a 2048Wh portable power station, which supports up to three hours of continuous capture.

We employ five OptiTrack Prime 13W MoCap cameras for tracking the IR-reflective markers fixed to the top of the user-worn headset; the set of markers are tracked altogether as a rigid body. 
We use a custom solution to calibrate the MoCap system, the rig cameras, and the headset cameras all into a shared 3D space. Then, the per-frame extrinsics of the headset cameras can be  inferred via the 6DoF transform of the tracked rigid body, such that the 3D ground truth computed from rig cameras can be precisely projected into the headset cameras for every frame.

All of the monochrome fisheye cameras run an autoexposure policy in real-time, to adapt to variable lighting conditions. Headset cameras are controlled by Quest 3's on-device autoexposure algorithm, while for the rig cameras, our capture software enforces a target average intensity of 60 over the entire image.

\begin{figure}[t]
    \centering
    \includegraphics[width=.9\linewidth]{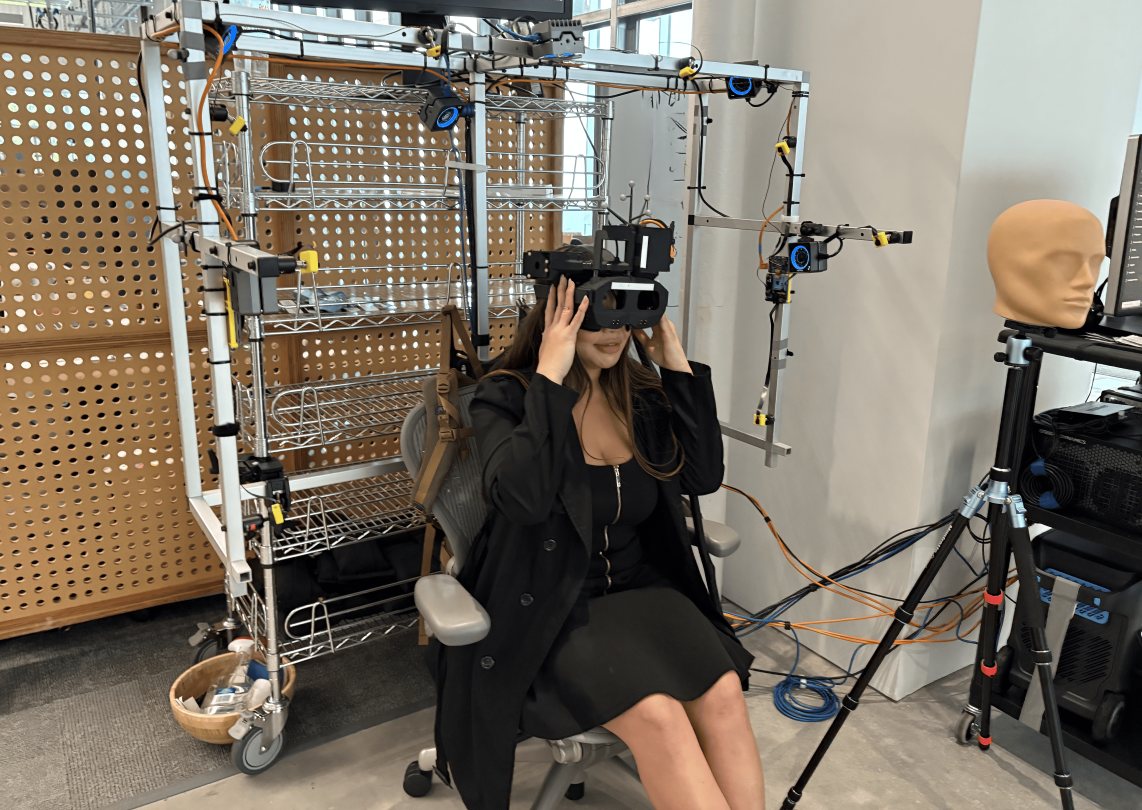}
    \caption{\textbf{Stationary configuration for inclusive capture.} The rig can be mounted statically to allow for seated data collection. This setup eliminates the weight burden of the backpack, accommodating participants with limited mobility or physical endurance.}
    \label{fig:sitting_setup}
\end{figure}

\subsection{Inclusive Design via Stationary Configuration}

To ensure that our dataset captures a diverse range of participants, we designed our system to be adaptable to different physical needs. While the primary configuration is a wearable backpack, the rig can also be fixed to a rolling cart, with a chair placed inside so the participant assumes the same relative position as if wearing it (see Figure~\ref{fig:sitting_setup}).

This configuration allows participants to perform hand-object interactions while seated, completely removing the physical burden of carrying the backstrap rig on the body. This adaptation is critical for inclusivity, as it enables participation from individuals with limited physical endurance, mobility constraints, or those who would otherwise be unable to support the weight of the rig for extended periods. This way, we ensure that our data collection is accessible to, and representative of, a broader demographic.

\section{SHOW3D Dataset Details}

\begin{figure}[t]
    \centering
    \includegraphics[width=\linewidth]{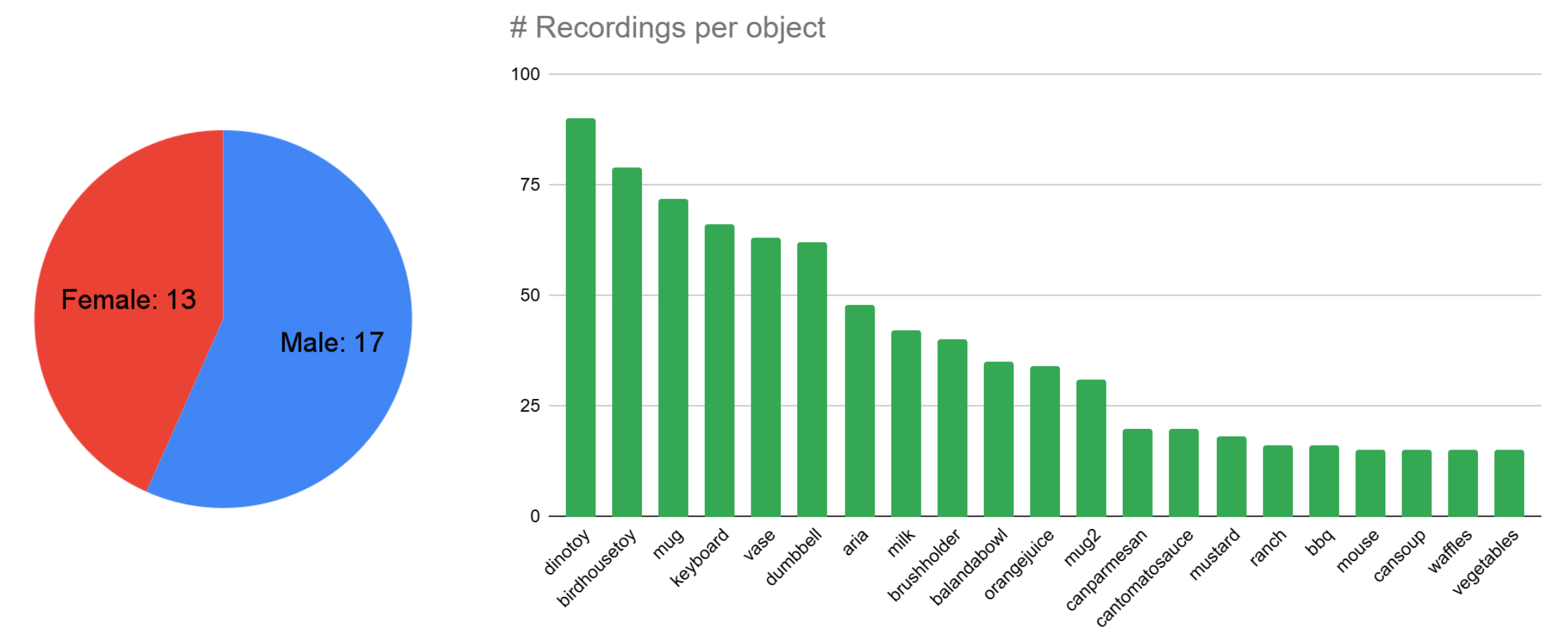}
    \caption{\textbf{SHOW3D dataset statistics.} 
    Left: participant distribution. 
    Right: number of recordings per each of the 21 objects.
    }
    \label{fig:dataset_stats}
\end{figure}

\customparagraph{Dataset Construction and Statistics.} 
We collected around 20 hours of synchronized ego-exo video data at 60Hz, resulting in 4.3M unique frames, or 43M unique images from all cameras. 
The dataset captures a large variety of indoor/outdoor environments, hand motions, and object interactions.
In a subset of around 2M frames, we captured hand-object interaction with a total of 21 objects procured from the list used in HOT3D~\cite{banerjee2025hot3d}.
The rest of the dataset feature bare-hand motion, hand-hand interaction, or hand-object interaction with other non-HOT3D objects that we currently do not have 3D scans for.

We recruited 38 human participants (24 male, 14 female) with a wide range of hand shapes and sizes.
All participants signed a form stating their consent for their collected data to be used for research purposes.
In addition, we employ a commercially available high-resolution optical scanning system to obtain a personalized hand model for each participant. The personalized model is a linear blend skinning mesh in UmeTrack~\cite{han2022umetrack} format, which can be fit to detected 3D keypoints via Inverse Kinematics.

\begin{figure}[t]
    \centering
    \includegraphics[width=\linewidth]{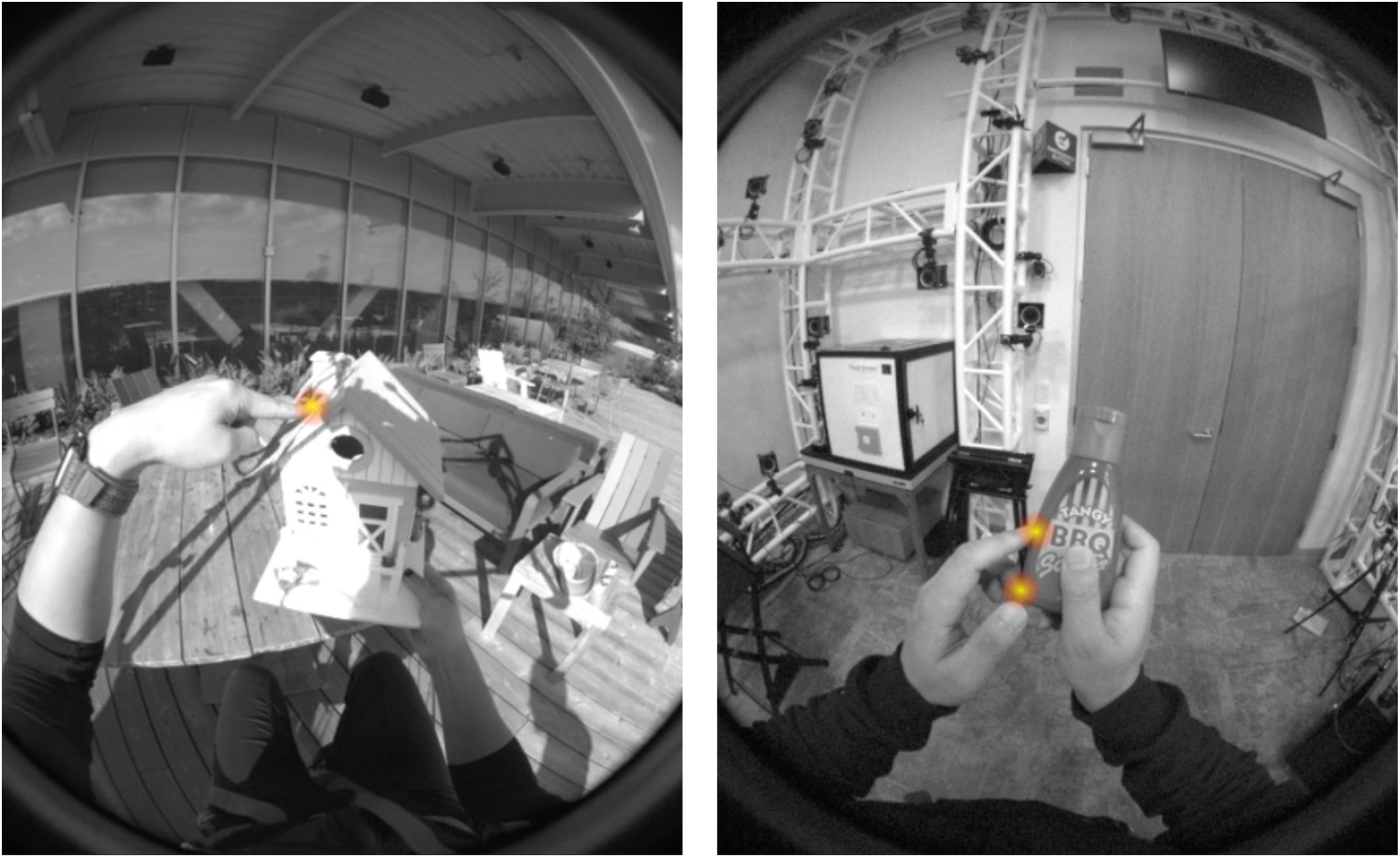}
    \caption{\textbf{Examples of contact annotations.}  Egocentric frames from SHOW3D with annotated contact regions (yellow). For visual clarity, contact annotations are only shown between the left hand and the manipulated object.}
    \label{fig:left_contact}
\end{figure}

\customparagraph{2D Segmentation Masks.} 
We automatically generate 2D segmentation masks for both hands and objects by projecting the annotated 3D meshes into each camera view using the calibrated intrinsic parameters. We can further utilize the projected mask as a prompt for the Segment Anything Model (SAM 2)~\cite{ravi2024sam} to obtain high-quality, refined 2D segmentation masks suitable for training 2D vision tasks.

\customparagraph{Contact Regions.} 
Leveraging our precise mesh reconstructions, we generate detailed contact annotations by calculating the Euclidean distance from every vertex on the hand mesh to the nearest point on the object surface. Vertices with a distance below some threshold (e.g., 5mm) are labeled as contact points. This results in continuous, vertex-level contact maps (see Figure~\ref{fig:left_contact}) that capture the functional grasp regions during dynamic interactions.

\section{3D Hand Annotation Quality}
\begin{figure}[t]
    \centering
    \includegraphics[width=0.9\linewidth]{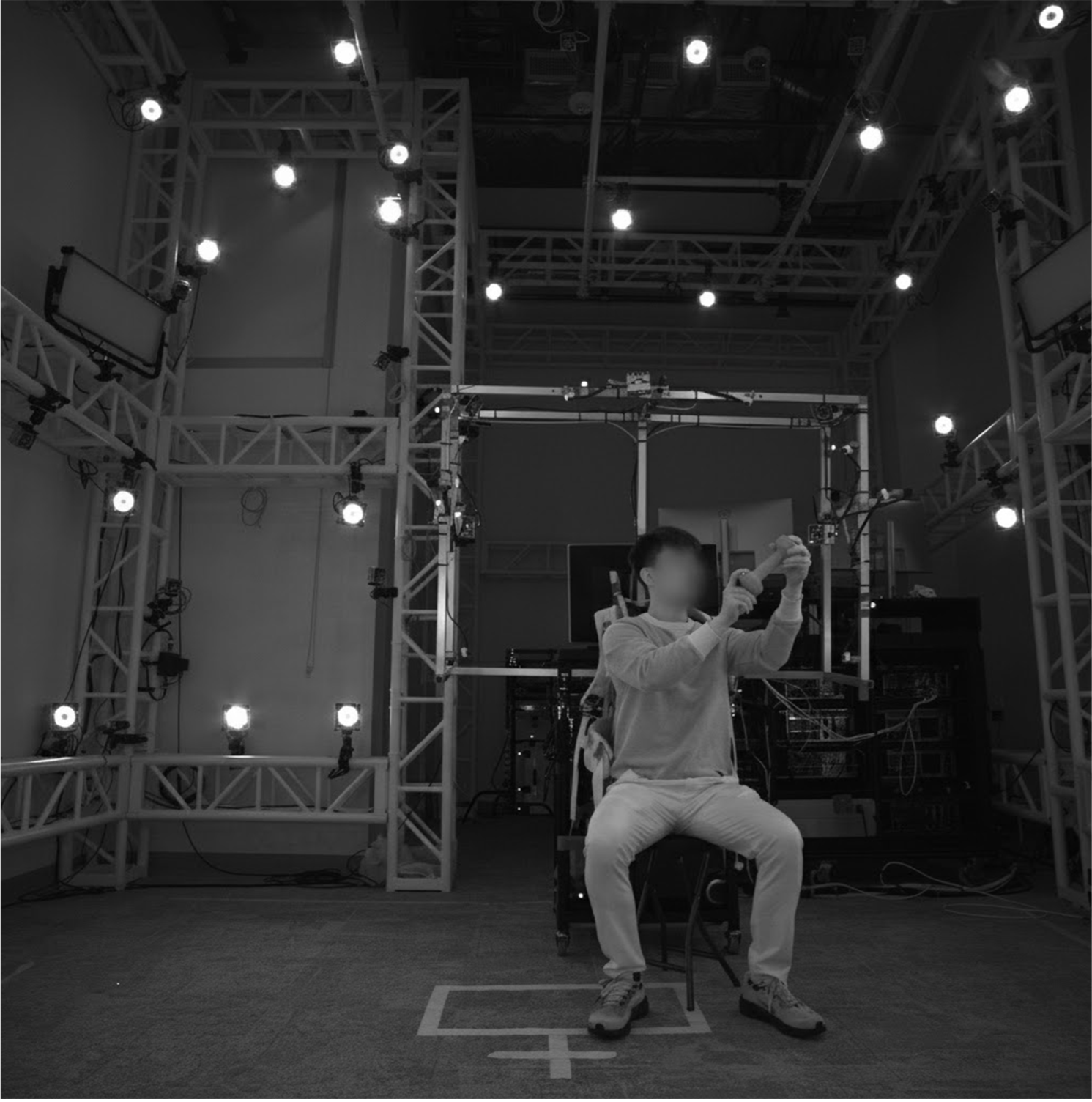}
    \caption{\textbf{Simultaneous rig--dome capture for ground truth validation.} 
    To evaluate the accuracy of our mobile annotation pipeline, we captured sequences where the subject wears our mobile rig inside a high-fidelity stationary multi-camera dome. 
    This allows for a direct comparison between our mobile-rig-derived annotations and the dome-derived ``gold standard.''}
    \label{fig:dome_capture}
\end{figure}

To validate the accuracy of the 3D hand poses generated by our mobile rig via our ego-exo pipeline (see Figure 3 of the main paper, and Figure~\ref{fig:hand_gt_pipeline}), we compare our automated annotations against two distinct ``gold standard'' sources: a stationary multi-camera studio system and human manual annotation.

\customparagraph{Experimental Setup.}
We define the two evaluation protocols as follows:
\begin{itemize}
    \item \textbf{Dome} (in-studio validation): We engaged a participant to wear our mobile rig \emph{inside} a large-scale, stationary multi-camera dome system equipped with 30 calibrated cameras (see Figure~\ref{fig:dome_capture}). The Dome system provides high-precision automated 3D keypoints via massive multi-view triangulation, serving as a proxy for human-annotated ground truth in a controlled environment. 
    To align the coordinate systems, we attached reflective markers to the mobile rig and utilized the dome's infrared motion capture system to calibrate the rig into the dome's 3D space.
    All data feeds—from both the rig and dome cameras—are synchronized with high precision.
    \item \textbf{Manual} (in-the-wild validation): To validate performance in realistic outdoor conditions where the Dome is unavailable, we sampled a subset of frames from our in-the-wild captures. Crowd-sourced annotators manually labeled 2D keypoints on all available camera views. We then obtained 3D ground truth via robust RANSAC-based triangulation of these manual 2D labels.
\end{itemize}

\begin{table}[t]
\centering
\small
\resizebox{\columnwidth}{!}{%
\begin{tabular}{c l c c}
\toprule
& & \multicolumn{2}{c}{\textbf{MPJPE (mm)}} \\
\cmidrule(lr){3-4}
\textbf{Reference} & \textbf{Category} & \textbf{Median}$\downarrow$ & \textbf{P90}$\downarrow$ \\
\midrule
\multirow{3}{*}{Dome}
  & No interactions          & 6.28 & 7.31 \\
  & Hand–hand interaction    & 7.46 & 8.95 \\
  & Hand–object interaction  & 7.90 & 12.53 \\
\midrule
\multirow{3}{*}{Manual (in the wild)}
  & No interactions          & 5.65 & 12.35 \\
  & Hand–hand interaction    & 6.03 & 14.63 \\
  & Hand–object interaction  & 6.36 & 16.48 \\
\bottomrule
\end{tabular}
}
\caption{Quantitative evaluation of our ground-truth hand annotations against two references: a high-coverage 30-camera dome (\textbf{Dome}) and manual annotations (\textbf{Manual}). We report median and 90th percentile per-frame MPJPE (mm).}
\label{tab:gt_evaluation}
\end{table}

\begin{figure}[t]
    \centering
    \includegraphics[width=\linewidth]{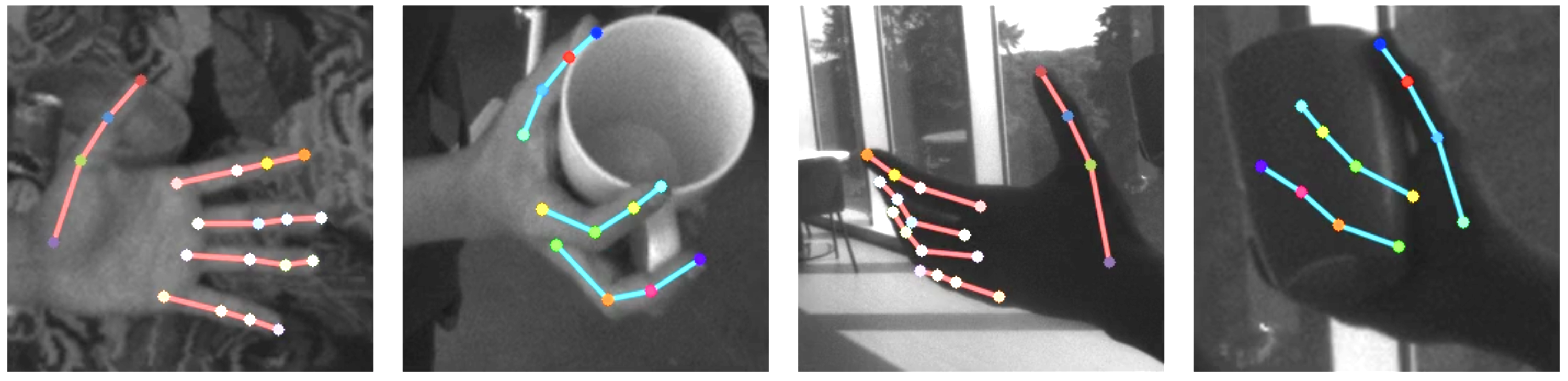}
    \caption{\textbf{Examples of manually annotated hand keypoints.} 
    Orange: left hand, blue: right hand. Note that the annotators are instructed to annotate only the keypoints that they have high confidence for.
    During evaluation, we only evaluate against annotated keypoints that have valid 3D triangulation.
    }
    \label{fig:manual_annotation}
\end{figure}

\customparagraph{Results and Analysis.}
Table~\ref{tab:gt_evaluation} reports the median and 90th percentile (P90) Mean Per-Joint Position Error (MPJPE) across three interaction complexity levels.

We observe that the \textbf{Manual} reference yields a lower median error than the \textbf{Dome} (e.g., 6.36mm vs. 7.90mm for hand-object interactions), but a higher P90 error (16.48mm vs. 12.53mm). 
We attribute this P90 discrepancy to the independence of the ground-truth generation methods. The Dome system relies on a multi-view algorithmic pipeline (automated keypoint detection and triangulation) that is methodologically similar to our own ego-exo pipeline. Consequently, the Dome and our system likely share similar failure modes—such as occlusion patterns or challenging lighting conditions—resulting in correlated errors that artificially dampen the P90 values. 
In contrast, human annotators provide a truly independent ground truth. They do not suffer from algorithmic biases and can accurately label challenging views that automated detectors might miss or misinterpret. Therefore, the Manual evaluation is a more rigorous ``stress test,'' exposing the true outlier failure cases of our triangulation-based approach that remain hidden when comparing against another automated system.

Importantly, across both evaluation protocols and all interaction types, our pipeline consistently achieves \textbf{sub-centimeter median accuracy}. Even in the most challenging scenario—hand-object interaction—our median error remains below 8mm against the Dome and below 7mm against manual annotation. 
These results confirm that our mobile, markerless capture system generates 3D ground truth of sufficient quality to train downstream egocentric perception models~\cite{han2020megatrack, han2022umetrack, pavlakos2024reconstructing, zhang2016learning, yang2024mlphand, prakash20243d, fan2023arctic}, successfully bridging the gap between studio fidelity and in-the-wild diversity.

\begin{figure*}[ht]
    \centering
    \begin{minipage}[b]{0.672\textwidth}
        \begin{subfigure}[b]{\textwidth}
            \centering
            \includegraphics[width=\textwidth]{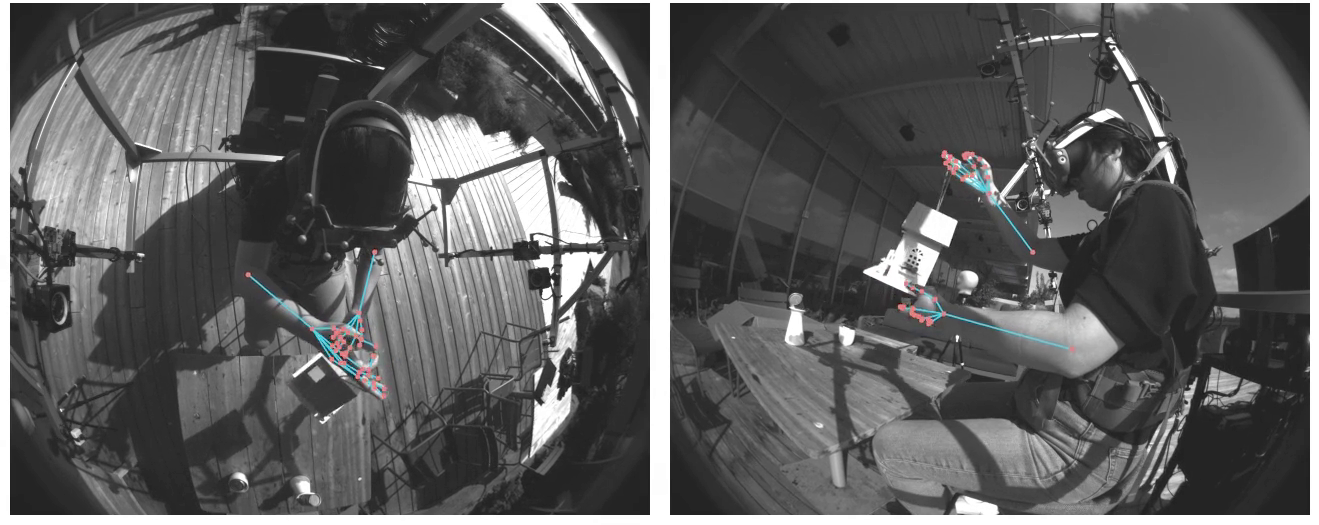}
            \caption{Triangulated Sapiens keypoints: hand joints and forearm.}
            \label{fig:subfig1}
        \end{subfigure}
        \vspace{0.02\textwidth}
        \begin{subfigure}[b]{\textwidth}
            \centering
            \includegraphics[width=\textwidth]{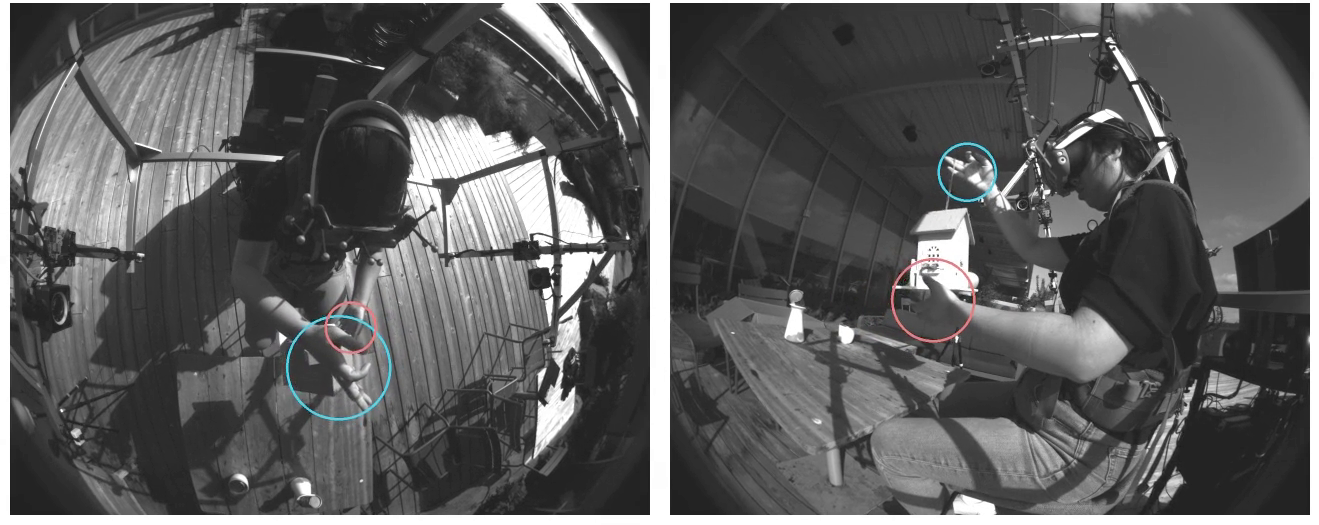}
            \caption{Inferred crop regions for both hands.}
            \label{fig:subfig2}
        \end{subfigure}
    \end{minipage}
    \hfill
    \begin{minipage}[b]{0.285\textwidth}
        \begin{subfigure}[b]{\textwidth}
            \centering
            \includegraphics[width=\textwidth]{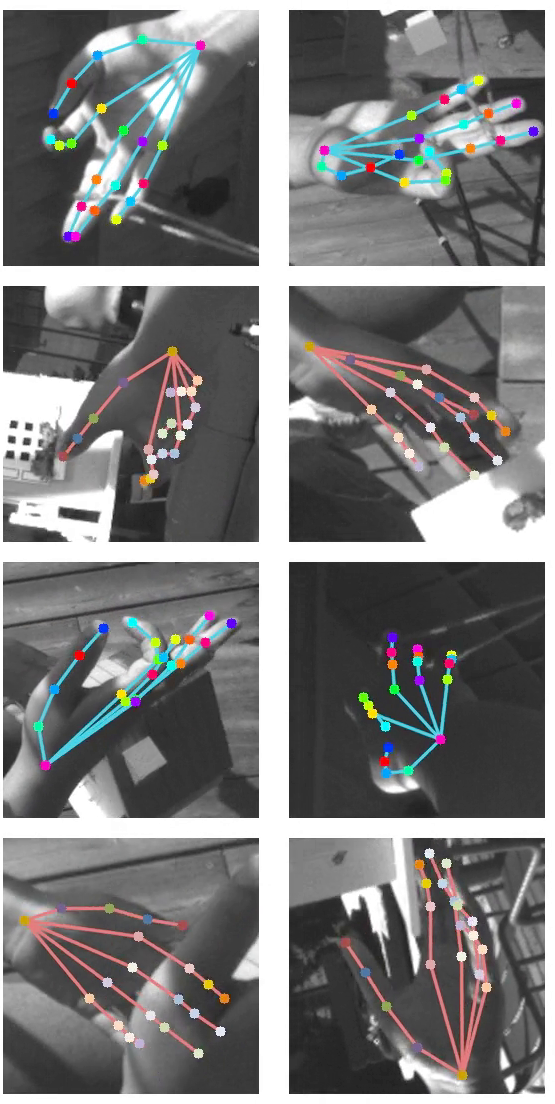}
            \caption{Final hand keypoint detections. Orange: left hand, Blue: right hand.}
            \label{fig:subfig3}
        \end{subfigure}
        \vspace{-15pt}
    \end{minipage}
    
    \caption{Our automated pipeline for 3D hand keypoint annotation. (a) Sapiens keypoints: we use the triangulated hand and forearm keypoints to define crop regions for hands. 
    (b) Crop regions are used to generate 256x256 perspective crops \cite{han2022umetrack} for both hands.
    (c) We apply InterNet~\cite{Moon_2020_ECCV_InterHand2.6M} on the perspective crops, and fuse them with the initial Sapiens detections to obtain final 3D hand keypoints.
    }
    \label{fig:hand_gt_pipeline}
\end{figure*}

\section{6DoF Object Annotation Quality}

To quantitatively evaluate our object pose annotations, we attach optical markers to 5 selected objects and track them using a MoCap system while they are being manipulated by hands.
We evaluate our GT object poses against the MoCap object poses (``gold standard'') in Table~\ref{tab:obj_annotation} by comparing the yield (\% of annotated frames), translation error (TE), and rotation error (RE).
The yield and accuracy of our GT object poses are generally very high, except for {\small \texttt{bottle\_bbq}}, which is a smaller and thus harder to detect object.
These results make SHOW3D the first HOI dataset to quantitatively evaluate both its 3D hand pose and 6DoF object pose annotations against independent references.
\begin{table}[h]
\footnotesize
\centering
\vspace{-6pt}
\resizebox{\columnwidth}{!}{%
\begin{tabular}{c c c c}
\toprule
Object & Yield (\%) & P50 TE (mm)$\downarrow$ & P50 RE (deg)$\downarrow$ \\
\midrule
birdhouse\_toy  & 89.5 & 3.51 & 1.73 \\
carton\_oj  & 92.6 & 3.20 & 1.23 \\
dumbbell\_5lb  & 86.1 & 1.53 & 1.63 \\
aria\_small  & 77.3 & 2.79 & 1.81 \\
bottle\_bbq  & 53.5 & 2.60 & 5.77 \\
\bottomrule
\end{tabular}
\vspace{-40pt}
}
\caption{Quantitative evaluation of our ground-truth object annotations against MoCap-derived poses as an independent reference.}
\label{tab:obj_annotation}
\end{table}

\section{Ground Truth Pipeline Details}
As mentioned in Section~3.2 of the main paper, we use a multi-stage ego-exo pipeline to obtain 3D hand annotations. 
We illustrate this pipline in detail with Figure~\ref{fig:hand_gt_pipeline}.

While the Sapiens~\cite{khirodkar2024sapiens} keypoint model exhibits strong generalization on in-the-wild images, we observe that its hand keypoint quality is often suboptimal, likely because the backbone's feature resolution is insufficient for recognizing finer details. 
Our insight is that applying a dedicated hand keypoint detector on cropped and resized images helps to improve performance; note that cropping can remove background variations to a large extent, resulting in an easier detection problem.
On the other hand, the initial Sapiens keypoint detections are robust enough for defining rough crop regions for both hands, even under heavy occlusion during object interaction.

\begin{table*}[!t]
\centering
\resizebox{\textwidth}{!}{%
\begin{tabular}{c l *{13}{c} c}
\toprule
\shortstack{Frames\\ahead} & &
\rotatebox{45}{pouring in} & \rotatebox{45}{picking up} & \rotatebox{45}{pouring out} &
\rotatebox{45}{opening} & \rotatebox{45}{stirring} & \rotatebox{45}{placing down} &
\rotatebox{45}{raising} & \rotatebox{45}{shaking} & \rotatebox{45}{tapping} &
\rotatebox{45}{washing} & \rotatebox{45}{checking} & \rotatebox{45}{inspecting} &
\rotatebox{45}{moving} &
Mean \\
\midrule

\multirow{3}{*}{30} & w/o text
 & 45.8 & 35.7 & 56.1 & 30.4 & 49.9 & 38.2 & 31.0 & 33.8 & 29.3 & 64.5 & 46.8 & 61.9 & \textbf{56.2}
 & 44.6 \\
 & w/ text
 & \textbf{28.3} & \textbf{23.6} & \textbf{37.3} & \textbf{20.5} & \textbf{35.0} & \textbf{29.4} & \textbf{24.5} & \textbf{27.5} & \textbf{26.1} & \textbf{59.7} & \textbf{44.9} & \textbf{61.2} & 59.1
 & \textbf{36.7} \\
 \hdashline \\[-2ex]
 & Improv.
 & 38.2\% & 33.9\% & 33.5\% & 32.6\% & 29.9\% & 23.0\% & 21.0\% & 18.6\% & 10.9\% & 7.4\% & 4.1\% & 1.1\% & -5.2\%
 & 17.7\% \\
\midrule

\multirow{3}{*}{60} & w/o text
 & 53.1 & 41.8 & 60.6 & 34.4 & 48.4 & 40.1 & 37.8 & 40.9 & \textbf{29.9} & \textbf{71.6} & 55.7 & \textbf{63.8} & 69.7
 & 49.8 \\
 & w/ text
 & \textbf{35.4} & \textbf{26.7} & \textbf{42.8} & \textbf{22.6} & \textbf{40.2} & \textbf{31.9} & \textbf{25.7} & \textbf{30.7} & 32.2 & 73.6 & \textbf{51.2} & 63.9 & \textbf{68.6}
 & \textbf{42.0} \\
 \hdashline \\[-2ex]
 & Improv.
 & 33.3\% & 36.1\% & 29.4\% & 34.3\% & 16.9\% & 20.4\% & 32.0\% & 24.9\% & -7.7\% & -2.8\% & 8.1\% & -0.2\% & 1.6\%
 & 15.8\% \\
\bottomrule
\end{tabular}%
}
\caption{Per-verb evaluation of text-driven 6DoF object pose forecasting, for a selected set of verbs. 
Reported are translation errors (mm, $\downarrow$) and the relative percentage improvement provided by text conditioning. 
The results are sorted by improvement at 30 frames. 
Our method shows robust gains on predictable actions (\emph{pouring}, \emph{opening}), but has limited impact on ambiguous ones (\emph{moving}, \emph{inspecting}).}
\label{tab:interaction_field_verb}
\end{table*}

As described in the main paper, we fuse hand keypoint detections from Sapiens and InterNet~\cite{Moon_2020_ECCV_InterHand2.6M} via robust triangulation.
We assign  a real-valued confidence to each triangulated 3D keypoint: this confidence value can be used as  a weight on positional constraints in the subsequent Inverse Kinematics (IK) optimization, to adjust the contributions from individual keypoints.
Our confidence formulation, intuitively, rewards keypoints for having a larger inlier count and a lower reprojection error.
Specifically, given RANSAC inlier count threshold $i_\textrm{T}$ and 2D reprojection error threshold $e_\textrm{T}$, and a triangulated keypoint with observed inlier count $i$ and average reprojection error $e_\textrm{rep}$ over inlier views,
we define the confidence $C$ as follows:
\begin{align}
E_\textrm{rep} & = \frac{e_\textrm{T}-e_{\textrm{rep}}}{e_\textrm{T}}\in (0,1),\\
 C  & = \left(E_{\textrm{rep}}\right) ^{{\gamma}/{max\{1,i-i_\textrm{T}\}}}.
\end{align}

First, the fraction $E_\textrm{rep}$ is inversely related to the average reprojection error, and bounded within $(0,1)$.
Then, we reward higher inlier counts by raising $E_\textrm{rep}$ to a power smaller than 1, thus increasing the resulting confidence value towards 1.
The reward factor can be additionally controlled by a free parameter $\gamma>0$.

Finally, we assign a per-frame confidence value to each hand after solving IK. 
In downstream experiments, we threshold this per-hand confidence to determine valid frames; the threshold can vary depending on the task.
We use a Bayesian formulation to compute the per-hand confidence, based on the intuition that the final error in the hand pose is compounded from the errors in both keypoint detection and IK.
For each  hand, we simply take the product between 1) the average confidence of its associated keypoints, and 2) a term inversely proportional to the residual error from IK solving.

\section{Additional Results on Text-driven\\6DoF Object Pose Forecasting}

In Table 3 of the main paper, we demonstrate that text conditioning improves forecasting accuracy by 29\% and 25\% on average across all object categories for predicting 30 and 60 frames ahead, respectively. However, aggregating results by object potentially ignores the underlying mechanism of \emph{how} language aids forecasting. We hypothesize that the ``kinematic utility'' of a text caption is less dependent on the object itself (e.g., \emph{dumbbell} vs. \emph{keyboard}) and more on the specific interaction protocol (e.g., \emph{pouring} vs. \emph{inspecting}).

To investigate this, we reorganize our evaluation based on the interaction verb (identified and extracted from the full text description) rather than the object class. Table~\ref{tab:interaction_field_verb} presents the translation errors grouped by a subset of 13 distinct verbs present in SHOW3D text descriptions, sorted by the relative improvement provided by text conditioning in forecasting 6DoF object pose at the 30-frame horizon.

\begin{figure*}
  \begin{center}
    \captionsetup{type=figure}
    \includegraphics[width=.88\textwidth]{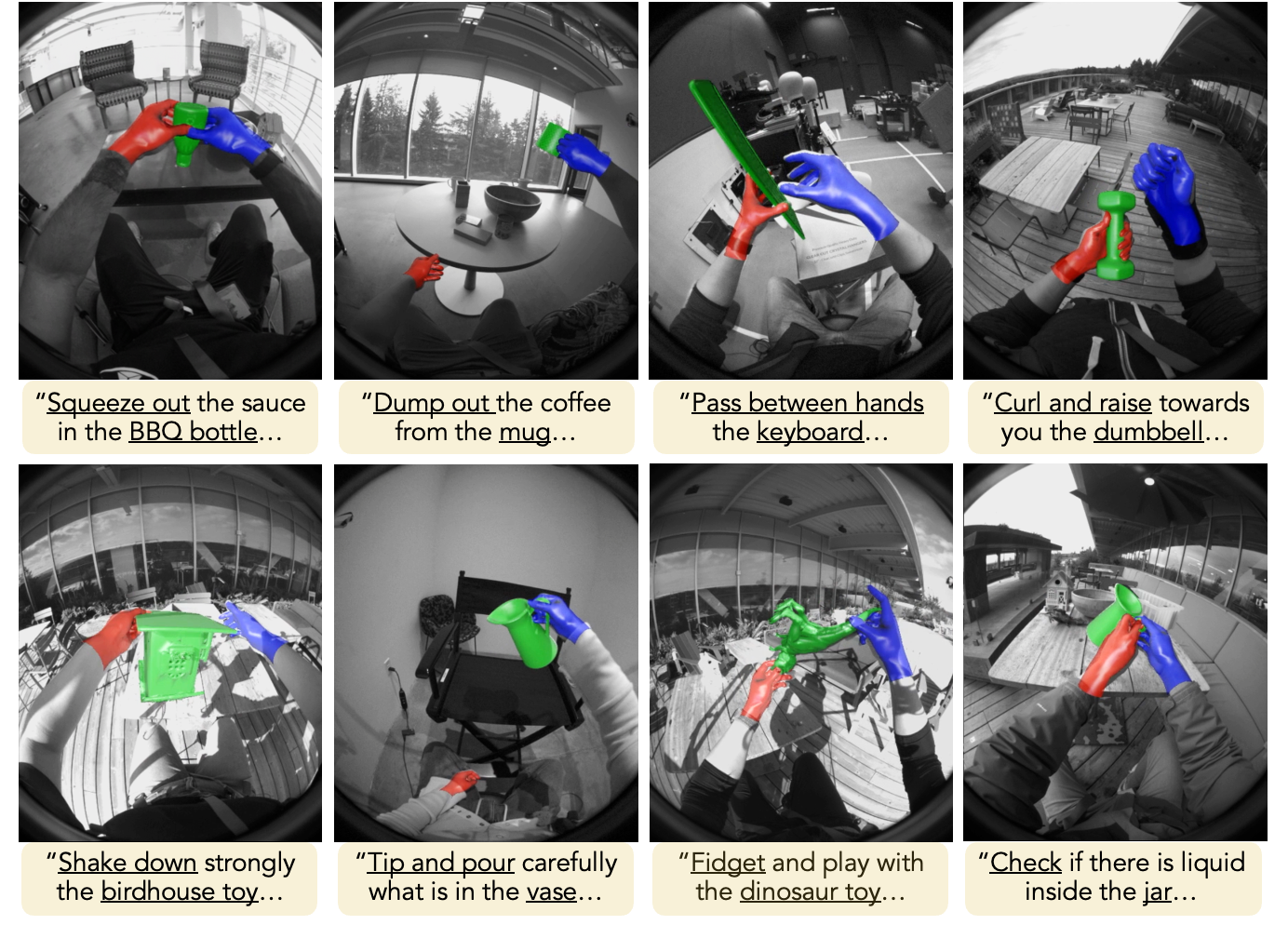}%
    
    \includegraphics[width=.88\textwidth]{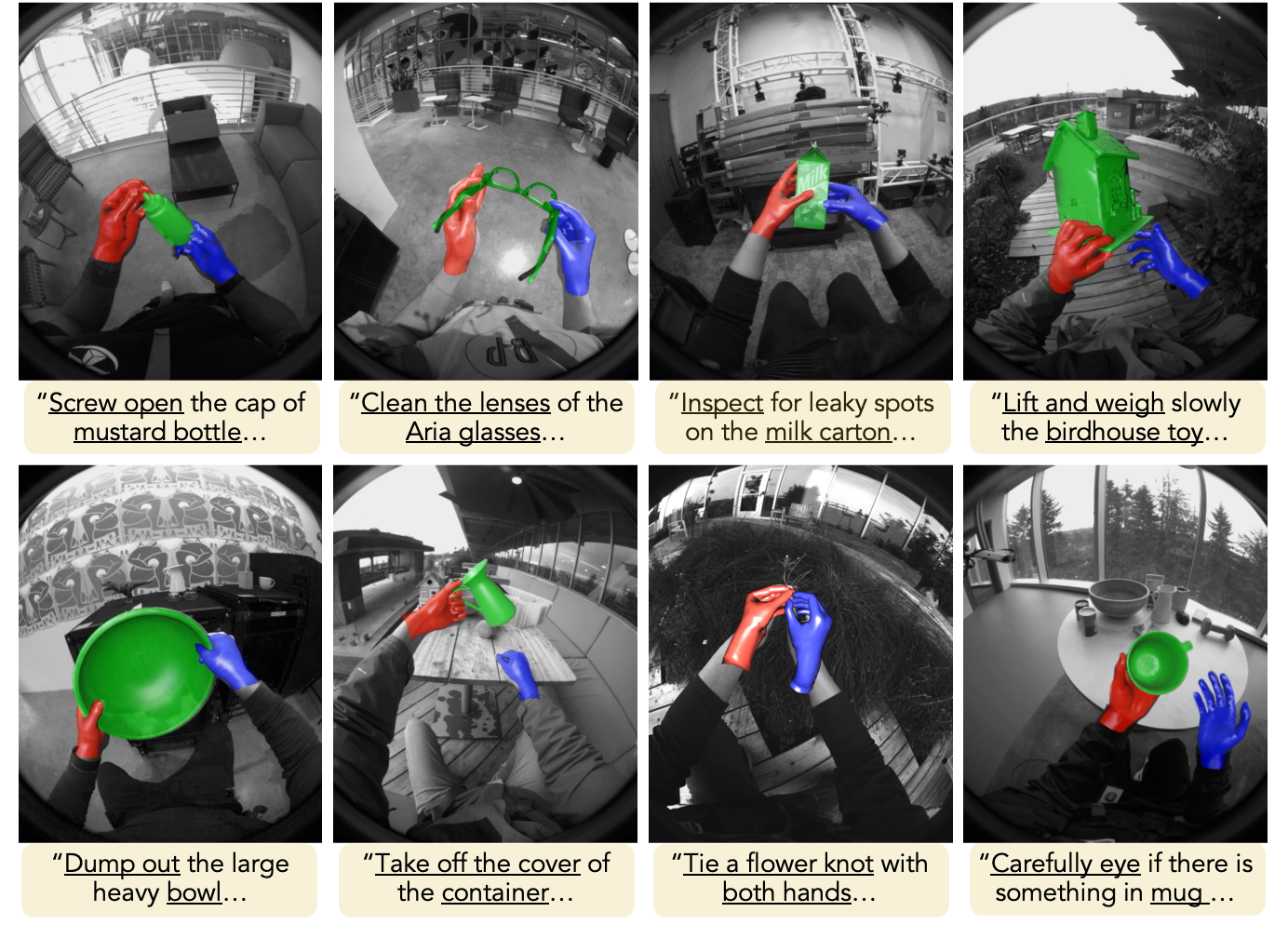}%
    \vspace{-4pt}
    \captionof{figure}{
    Additional ground truth examples from SHOW3D: hand pose (red and blue), object pose (green), and text captions. 
    }
    \label{fig:suppmat_good1}
  \end{center}%
  \vspace{6pt}
\end{figure*}

\begin{figure}[t]
    \centering
    \includegraphics[width=\linewidth]{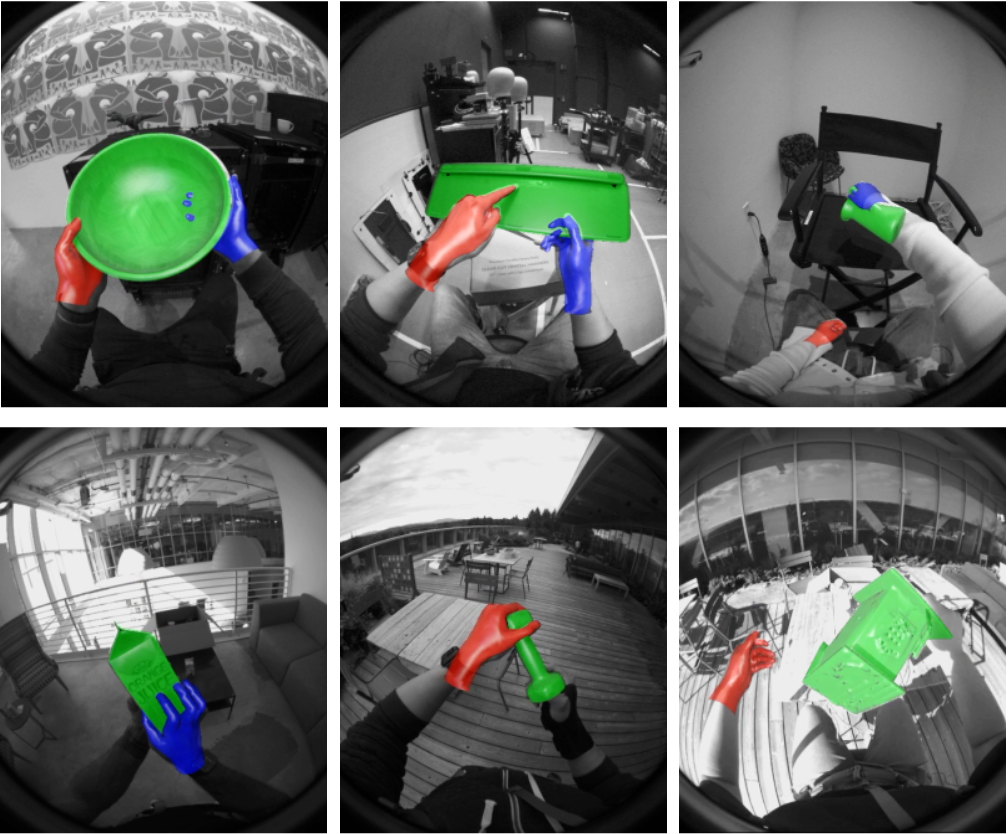}
    \caption{\textbf{Hand annotation failure cases.} We visualize two common error modes in our automated annotation pipeline. 
    \textbf{Top row:} Examples of \textbf{inaccurate 3D pose estimation}. While all entities are successfully detected, minor errors in the estimated 6DoF pose result in physically implausible interpenetrations (clipping) between the hand and object meshes.
    \textbf{Bottom row:} Examples of \textbf{missed hand detections}. In scenarios with severe occlusion—where a hand is significantly obstructed by the object or the opposing hand—our pipeline occasionally fails to detect the occluded hand entirely.}
    \label{fig:failure_cases_general}
\end{figure}

\begin{figure}[t]
    \centering
    \includegraphics[width=\linewidth]{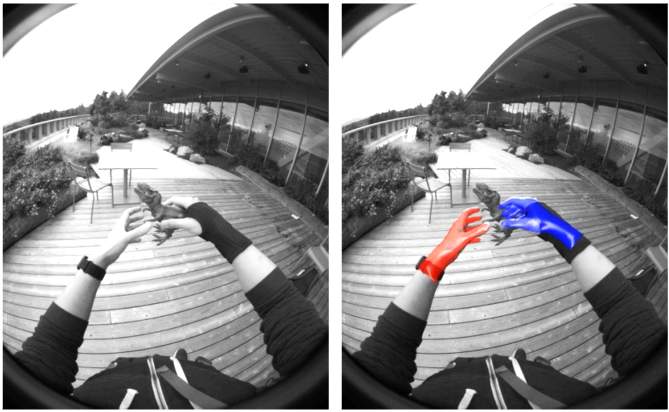}
    \caption{\textbf{Object annotation failure case:} Missed object under severe occlusion and outdoor illumination.
    \textbf{Left:} The raw image from egocentric view.
    \textbf{Right:} The projected 3D hand meshes (red/blue) overlaying the image.
    Although the hands are correctly tracked, the object is not detected due to occlusion from both hands and strong outdoor lighting, and thus is entirely not tracked.}
    \label{fig:failure_noobj}
\end{figure}

\begin{figure}[t]
    \centering
    \includegraphics[width=\linewidth]{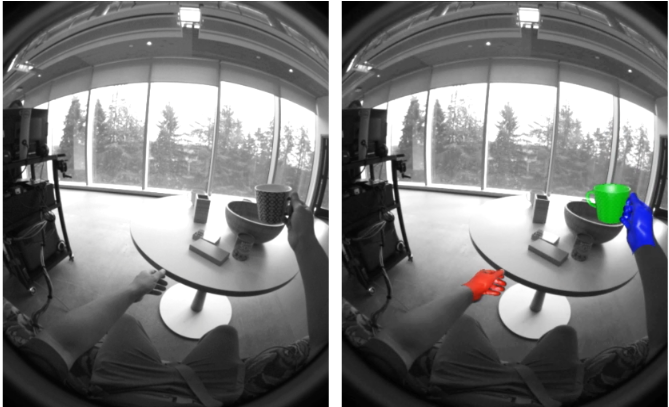}
    \caption{\textbf{Object annotation failure case:} Rotational ambiguity in symmetric objects.
    \textbf{Left:} The raw image from egocentric view. 
    \textbf{Right:} The projected 3D object mesh (green) overlaying the image. 
    Note that the pipeline correctly tracks the 3D position but fails to resolve the azimuthal rotation, incorrectly predicting the handle to be visible (``hallucinated'') despite not being present in the image.}
    \label{fig:failure_hallucination}
\end{figure}

\customparagraph{Impact of Action Predictability.}
The results reveal a strong correlation between the \textit{predictability} of an action and the performance gain from text conditioning. 

We observe the largest improvements in complex actions that involve specific, structured 3D state changes, such as \emph{pouring in} (38.2\% improvement), \emph{picking up} (33.9\%), and \emph{opening} (32.6\%). In these scenarios, the visual history alone may be ambiguous; for example, a hand approaching a bottle could precede lifting, opening, or sliding. The text caption serves as a strong prior that disambiguates the intent, collapsing the multimodal future distribution into a specific trajectory (e.g., the typical arc that the hand and object usually take to pour).

Conversely, actions characterized by high ambiguity or stochasticity benefit least from having text description as an additional input. In the task of predicting 6DoF object pose 30 frames ahead, \emph{inspecting} (1.1\%) and \emph{moving} (-5.2\%) show negligible or even negative differences. We attribute this to the wide variability in how users execute these actions; knowing that a user is ``inspecting'' an object predicts that the object will likely be rotated, but it does not specify the axis or magnitude of that rotation. In these high-entropy cases, the text describes the \emph{state} of the interaction but provides little kinematic signal regarding the future trajectory of the hand and object in 3D space.

This breakdown shows that the value of text-conditioning in SHOW3D is not uniform across hand-object interactions. Language features act as a high-level control signal that is most effective when the semantic intent dictates a specific kinematic protocol, effectively narrowing the search space for the forecasting model.

\section{Additional Ground Truth Examples}

We provide additional visualizations of our 3D ground truth annotations in Figure~\ref{fig:suppmat_good1}. 
These examples further illustrate the diversity of our captured environments and the high fidelity of our markerless tracking pipeline. 

\textbf{Failure Modes.} To contextualize these examples, Figure~\ref{fig:failure_cases_general} summarizes the two most common failure types observed in our automated hand annotation pipeline. The top row illustrates cases where all entities are successfully detected, but slight inaccuracies in the estimated 6DoF hand pose lead to physically implausible interpenetrations with object meshes. The bottom row shows the opposite failure mode, in which a hand is missed entirely due to heavy occlusion from the object or the opposing hand. While both issues occur infrequently relative to the dataset scale, they underscore the inherent challenges of annotating fine-grained hand–object interactions in egocentric views.

We also observe specific failure modes of our object tracking pipeline. In Figure~\ref{fig:failure_noobj}, the hands are successfully tracked but the object is missed entirely due to severe occlusion and strong outdoor illumination. In these settings, our object tracking pipeline, which relies on DINOv2~\cite{oquab2023dinov2} features, struggles when object features are either heavily occluded or when strong sunlight induces appearance changes, causing the object to go undetected by some or all cameras even when it is partially visible.

A second failure mode is shown in Figure~\ref{fig:failure_hallucination}, where the object tracking pipeline accurately localizes the 3D position of the cup but fails to estimate its azimuthal rotation, resulting in the handle being “hallucinated’’ in a visible position (right, green mesh) despite being fully occluded in the input image (left). This likely occurs because the DINOv2 features used in our pipeline~\cite{nguyen2023cnos, ornek2024foundpose, nguyen2025gotrack} are highly semantic and robust to local variations. While this robustness aids in tracking the dominant cylindrical body under occlusion, the patch-based features may lack the spatial sensitivity required to infer the orientation of the thin handle, particularly when the object appears largely rotationally symmetric.

\end{document}

%% file: preamble.tex
\usepackage{url}

\usepackage{algorithm}
\usepackage{algorithmic}

\usepackage[utf8]{inputenc} 
\usepackage[T1]{fontenc}    
\usepackage{multirow}
\usepackage{tikz}

\usepackage{booktabs}
\usepackage[font=small]{caption}

\usepackage{url}            
\usepackage{booktabs}       
\usepackage{amsfonts}       
\usepackage{nicefrac}       
\usepackage{microtype}      
\usepackage{xcolor}         
\usepackage{amsmath}
\usepackage{pifont}
\usepackage{adjustbox}
\usepackage{graphicx}

\definecolor{PaleBlue}{RGB}{100, 160, 190}
\definecolor{Purple}{RGB}{153, 51, 255}

\usepackage{gensymb}
\usepackage{booktabs}
\usepackage{tabularx}
\usepackage{array}
\usepackage{multirow}
\usepackage{graphicx}
\usepackage{colortbl}
\usepackage{pifont}
\usepackage{arydshln}

\newcommand{\cmark}{\ding{51}}%
\newcommand{\xmark}{\ding{55}}%
\newcolumntype{Y}{>{\centering\arraybackslash}X}

\newcommand\customparagraph[1]{\vspace{0.7em}\noindent\textbf{#1}}

